\documentclass{article} % For LaTeX2e
\usepackage{iclr2027_conference,times}

\usepackage{amsmath,amsfonts,bm}

\def\eqref#1{equation~\ref{#1}}
\def\1{\bm{1}}

\DeclareMathAlphabet{\mathsfit}{\encodingdefault}{\sfdefault}{m}{sl}
\SetMathAlphabet{\mathsfit}{bold}{\encodingdefault}{\sfdefault}{bx}{n}

\usepackage{hyperref}
\usepackage{url}
\usepackage{booktabs}
\usepackage{graphicx}
\usepackage{makecell}
\usepackage[table]{xcolor}     % \cellcolor{red!30} — the [table] option loads colortbl
\usepackage{amsmath}
\usepackage{tabularx}
\usepackage{array}
\usepackage{multirow}
\usepackage{xltabular}
\usepackage{amsfonts}
\usepackage{nicefrac}
\usepackage{microtype}
\usepackage{wrapfig}
\usepackage{listings}
\usepackage{amssymb}

\newcommand{\benchmark}{ReLiveGym}
\newcommand{\benchmarkntask}{8}

\title{\benchmark{}: Evaluating Long-Lived Agents over Weeks of Replayed Reality}

\author{
Xisen Jin\textsuperscript{1,2},\;
Jingheng Li\textsuperscript{2},\;
Zhenglun Chen\textsuperscript{1},\;
Junyi Du\textsuperscript{1},\;
Xiang Ren\textsuperscript{1,2}
\\
\textsuperscript{1}Sahara AI\;
\textsuperscript{2}University of Southern California
}

\iclrfinalcopy % Uncomment for camera-ready version, but NOT for submission.
\begin{document}

\maketitle

\begin{abstract}
As large language model (LLM) agents become widely adopted, they are increasingly deployed for tasks that require persistent monitoring or recurring actions (e.g., market analysis).
These agents are expected to operate unattended for days or weeks, act at the right timing, and adapt to the dynamic environment over time.
These challenges are not fully captured in the existing long-horizon agent work, as they often consider a static environment that is not temporally changing.
We introduce \benchmark{}, a diagnostic evaluation environment of \textit{long-lived} tasks in which agents act sparsely over simulated weeks of chronologically replayed real-world news, market, and social-media streams. The tasks span diverse levels of time sensitivity, reasoning intensity, and recurrence.
Across eight base language models, we investigate how model choice and harness design affect agent performance on such long-lived tasks.
Our results show that how agents determine when to act arises as an important harness-design axis for long-lived tasks; and that the optimal design varies across tasks and sometimes model choices as well.
We also evaluate how continuous learning from hindsight feedback affects performance and addresses failure modes observed in these long-lived tasks.
These findings indicate model choice, action timing mechanism, and use of feedback as important considerations in the design of long-lived agents. Code: \url{https://github.com/SaharaLabsAI/ReLiveGym}.

\end{abstract}

\section{Introduction}
\label{sec:intro}

Large language model (LLM)-powered AI agents are increasingly deployed to operate unattended on tasks that unfold over days to months: from daily automation~\citep{OpenAI2026Automation, n8n2026} to monitoring news for market-moving events~\citep{Qian2025WhenAT}.
We refer to these systems as \textit{long-lived} agents: agents that act intermittently over days to months in an evolving world in pursuit of an instruction or goal. They must act at the right moment, adapt to naturally occurring changes in their environment, and ideally continuously improve over time as the task recurs.

%which differ from long-horizon or long-running agents, which act non-stop towards solving complicated programming, math, and science problems~\citep{chan2025mlebench, Novikov2025AlphaEvolveAC,Merrill2026TerminalBenchBA, Zhu2026EdgeBenchUS}. 
%Unlike long-horizon agents that execute many consecutive steps until accomplishments of goals~\citep{Novikov2025AlphaEvolveAC, Lu2026TowardsEA}, these \textit{long-lived agents} must allocate limited attention over time: waking too rarely risks missing short-lived opportunities, whereas waking too often wastes inference and information-query budgets. 
%Long deployments also expose agents to repeated outcomes and delayed feedback, creating opportunities for test-time improvement~\citep{wang2024voyager, Zhu2025ScalingTC, Lee2026RecursiveHS}. Long-lived agents must therefore decide not only what to do, but also when to observe and act and how to use experience from earlier decisions to improve later ones.

Existing works often capture only parts of the challenges in long-lived agents. Long-horizon or long-running agents~\citep{chan2025mlebench, Novikov2025AlphaEvolveAC,Merrill2026TerminalBenchBA, Zhu2026EdgeBenchUS} act non-stop towards solving complicated programming, math, frontier science and engineering problems; however, unlike long-lived agents, these works typically do not assume the environment evolves chronologically without agents' actions. Conversely, some recent works study an independently evolving environment as an important challenge in practical agent use cases. In this setup, 
GAIA2~\citep{froger2026gaia} evaluates agents in mobile app use and demonstrates that agents struggle with making timing decisions, and SentinelBench~\citep{Maldaner2026SentinelBenchAB} further compares different wait mechanism designs. However, their tasks span only minutes to hours; challenges and opportunities such as continuous improvement can emerge in a longer time span. Evaluations in live environments (\textit{e.g.}, real-stock-market trading~\citep{Qian2025WhenAT, Ma2025AgentTA}) or over replayed streams~\citep{Zhou2021TradeTE, Goel2026FutureSimRW, Chen2025StockBenchCL} may span weeks to months, but often invoke agents on a fixed hourly or daily schedule; the choice of timing policy is rarely evaluated. %Proactive agents~\citep{Lu2024ProactiveAS, yang2025contextagent} evaluate agent's capability to react to events use as user's computer use activity; we consider the case where agents can either trigged by time elapse, events, or schedules.

\begin{figure}
  \centering
  \includegraphics[width=\textwidth]{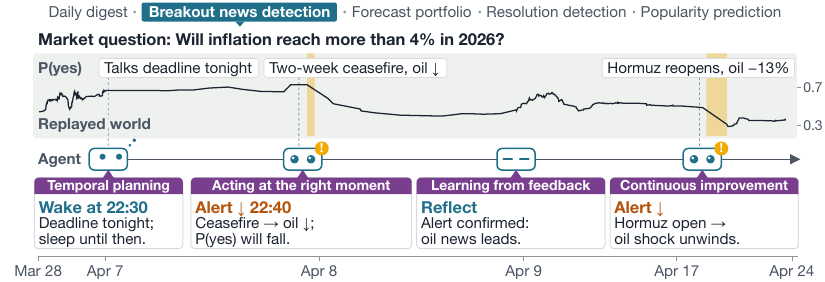}
  \caption{\textbf{A long-lived task in \benchmark{}.} The agent watches one replayed prediction market and news for four weeks and must alert before the price moves (shaded windows). It wakes sparsely, plans a wake for a known deadline, alerts before the market reacts, continously improve by reflecting on the confirmed outcome, and applies the lesson ten days later.}
  \label{fig:intro}
\end{figure}

To study the challenges faced by long-lived agents, we introduce \benchmark{}, a diagnostic evaluation environment comprising \benchmarkntask{} tasks that span weeks of chronologically replayed real-world streams. \benchmark{} replays news, social media, and market data on a simulated clock following~\citet{Andrews2025ARESU, Goel2026FutureSimRW}. Its tasks vary along several axes: some admit fixed-interval schedules, whereas others require timely, event-driven reactions.
Some tasks require strong forecasting capabilities, whereas others emphasize the reliable execution of routine actions. Some tasks provide delayed feedback, creating opportunities for continuous test-time improvement.
Each task is evaluated using deterministic metrics, such as accuracy and alert delay. Figure~\ref{fig:intro} illustrates an example breakout news detection task in \benchmark{} and the agent capabilities it requires.

Our key research questions are which factors affect the performance of long-lived agents and which failure modes arise empirically. We evaluate the effects of base LLM choice and multiple axes of harness design.
Across tasks, base-model rankings are fairly consistent: GPT-family models perform best on reasoning-intensive tasks, whereas Qwen models perform best on routine tasks.
We find that how an agent determines when to act arises as an important dimension of harness design; the effectiveness of each trigger mechanism depends on the task and, in some cases, the base model.
We further evaluate continuous improvement methods on tasks that provide hindsight feedback. We find that simple methods, such as parallel sampling of agent trajectories~\citep{koh2025tree, GonzalezPumariega2025ScalingAF}, mitigate empirically observed failure modes, including persistence with a poor policy chosen at the outset. The performance of a given base LLM can vary substantially across repeated runs, but trigger mechanisms and continuous-improvement methods can improve its stability. We expect the artifact and these findings to contribute to the development of long-lived agents.

%Our experiments reveal substantial variance in the performance of long-lived agents, not only across configurations but repeated runs of the same configuration, even on routine tasks. 
%Case studies reveal part of failure modes: \textit{late} failures, in which an agent deviates from its instructions after weeks of normal operation; \textit{early} failures, in which an agent commits to a poor strategy at the outset and never revisits it. 
%Harness design shifts both performance and cost, and the best trigger mechanism depends on the task and the model: \textit{e.g.,} a fixed schedule wins on daily digest, whereas customizable watchers are most helpful for some models on popularity prediction.
%Cost is not the mechanism: LLM spend and score are only weakly correlated. Hindsight feedback helps under identifiable conditions. 
%Reflective memory curated from hindsight feedback barely improves performance; whereas best-of-$N$ selection with pruning with hindsight feedback effectively identify and mitigates runs failing early. %Overall, the results imply that task-specific harness designs are crucial for optimizing cost, performance, and stability of long-lived agents.
%The results imply that there is not one harness that fits for all models and tasks; model and task-specific harness designs are crucial for performance and stability of long-lived agents.

In summary, our contributions are: (1) \benchmark{}, an evaluation environment for long-lived agents comprising \benchmarkntask{} tasks over weeks of chronologically replayed data; and (2) an analysis of how base-model choice and multiple axes of harness design affect long-lived agent performance, together with an empirical study of failure modes.

%We find that harnesses that allow LLMs to create and execute watcher code outperform those that invoke agents according to human-curated schedules, although they are often more costly. Test-time learning algorithms effectively acquire shallow knowledge, such as label distributions, but fall short of improving reasoning capabilities over time. The optimal harness depends on both the base model and the task.

% discard this for now
%Given that optimal harness designs vary, we ask whether an LLM can design a task-specific harness. Conceptually, the designing step is itself a part of the agent program. We notice that the resulting designs often make rigid timing decisions and omit test-time learning components such as memory; when implemented, they underperform human-curated harnesses. Overall, our benchmark reveals persistent challenges for existing agent harnesses and models on long-lived tasks.
\section{Problem Formulation} 
\label{sec:formulation}

We formalize long-lived agents as programs that act intermittently in an exogenously evolving environment. The formulation specifies the agent program, its timed interactions with the environment, and the task constraints under which it operates.

\paragraph{Agents.}
An agent is a \emph{program} $\mathcal{A} = (H, M)$ comprising a harness $H$ and a base LLM $M$. The harness is the code that defines prompts and tools, initiates new turns, and manages context and memory. Together, $H$ and $M$ induce a policy that maps the interaction history to the next action. Given an initial user instruction $x$, the agent acts at a discrete sequence of timestamps over the horizon $[0,T]$, taking its $i$-th action $a_i$ at time $t_i$.
These actions include querying the current environment state $E(t_i)$ and committing user-facing results $y_j$ at times $t_j$.
Agents incur costs when invoking LLMs and operate under a cost budget $B$. We calculate these costs using provider rates listed in Table~\ref{tab:cost-rates} in the Appendix and hold the rates fixed throughout the study.

\paragraph{Long-lived tasks and environments.} 
A long-lived task is defined by an environment $E$, an initial user instruction $x$, a cost budget $B$, a time horizon $T$, and an evaluation metric $f$.
The environment $E$ evolves continuously over a horizon of world time $[0, T]$ (days to weeks). Crucially, this evolution is \emph{exogenous}: the world state $E(t)$ evolves on the world's clock $t$ whether or not the agent is watching.
The metric function scores each commitment $\langle y_j, t_j \rangle$, so the value of a commitment depends on its timing $t_j$. We assume that the cost budget, time horizon, and evaluation metric are fixed and known to the agent at the outset and therefore need not be inferred; in practice, we include them in the agent's instruction. However, the agent cannot observe the future evolution of $E(t)$, which is grounded in changes recorded in the real world.

A solution to a \benchmark{} task is a \emph{program}. It may be a ReAct agent~\citep{yao2023react}, a ReAct agent invoked on a fixed cadence, a fixed LLM-based workflow, or a deterministic program that does not invoke an LLM. We focus our evaluation on the first two because a general ReAct loop can be transferred across tasks by changing its instruction and available tools.%, whereas workflows and deterministic programs typically require task-specific logic curated in advance.

\section{\benchmark{}: Evaluating Long-Lived Agents over Replayed Reality}

%The key objective of \benchmark{} is to evaluate agents on a chronologically replayable environment backed by the real world data. The tasks are drawn from three worlds (markets, news, and social media) and are chosen to span the capability requirements that distinguish tasks in different axes. We describe the task curation process and evaluated capalities.
\benchmark{} consists of \benchmarkntask{} tasks built over chronological replay of real-world data for weeks of world time. The tasks are chosen to span the capability requirements across different axes. We describe the worlds and tasks, their curation process, and the evaluated capability axes. We then introduce how the agents are evaluated in a simulated environment.

\subsection{Curating Tasks and Simulated Worlds for \benchmark{} Tasks}
\label{ssec:simu_world}
\benchmark{} tasks are created by first identifying replayable streaming data in the real world. We use timestamped news, market, financial-report, and social media data.

\paragraph{Simulated worlds.} 
We build a news world from Common Crawl's news crawl with full-text BM25 search APIs available for the agent. The publish time of an article is inferred from both the publisher's claimed time and the crawler's crawl time.
%The publish time is inferred from self-reported publish time and the crawler's time. 
We build the prediction-market world from the non-sports Polymarket markets with $\ge \$100k$ volume that overlap the window, replaying each market's question, open date, minute-resolution prices and resolution.
For market prices, we replay three exchanges' 5-minute candles.
For financial filings, we snapshot the SEC's public filing data\footnote{\url{https://data.sec.gov}} for every held company and serve it as of the clock. For social media, we replay Reddit posts with their content and timestamped comment trees. 

We sample data only after March 1, 2026, which is beyond the known knowledge cutoff dates of the evaluated LLMs, which minimizes the impact of data contamination. We release the data curation pipeline so that future LLMs can be evaluated on a newer environment.

\begin{figure}
  \centering
  \includegraphics[width=\textwidth]{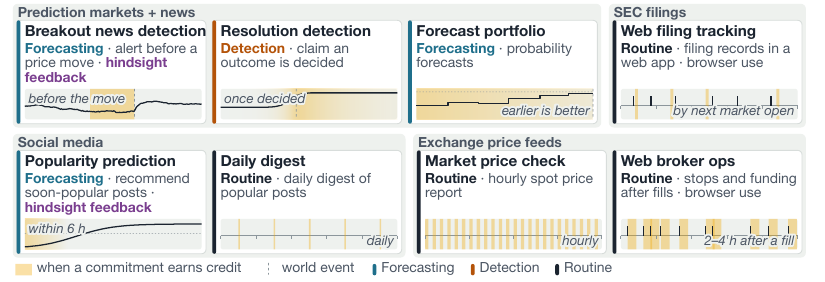}
  \caption{\textbf{The eight \benchmark{} tasks} curated over real-world replayed data streams. Agents are evaluated for their commitments such as breakout news alerts in the simulated weeks. The tasks diverge in challenge axes such as their timing requirements and reasoning depth.}
  \label{fig:tasks}
\end{figure}

\subsection{\benchmark{} Tasks and Capability Axes}
\label{ssec:capability_axes}

After the identification of replayable data, the authors proposed task templates that are realistic and cover diverse axes of the agent capability challenges.
The task templates are instantiated as concrete tasks (\textit{e.g.,} tracking concrete markets for a period) when grounded in a time window. We list \benchmark{} tasks across diverse axes of the agent capability in Figure~\ref{fig:tasks}, with details included in Table~\ref{tab:task-details} in the Appendix. Below we introduce the key challenges and the associated tasks.

\paragraph{Action timing and schedule.} \benchmark{} tasks diverge in how sharply the value of a commitment decays with earliness or delay, and whether a task implies a fixed schedule or is driven by world events. At one end, the task \textit{daily digest} implies a fixed schedule: a once-a-day wake-up is the right policy. In the middle, \textit{forecasting portfolio} is gradually time-sensitive: a forecast may be reviewed at any time before a question resolves, while earlier forecasts earn more credit. At the far end, tasks are sharply sensitive in one direction: \textit{breakout news detection} requires early alerts, and is worth credit only before a market price move; \textit{resolution detection}, in contrast, penalizes early claims before a controversy is settled in the world. \textit{Web broker operations} requires agents to act before a deadline that is several hours after an event happens. 

\paragraph{Reasoning depth and complexity.} \benchmark{} tasks diverge in their reasoning depth and complexity. Some tasks, such as \textit{price check} and \textit{daily digest}, primarily require applying deterministic rules; this also applies to tasks like~\textit{web filing} and~\textit{web broker operations}, which add complexity by requiring browser use. Conversely, some tasks require synthesizing evidence across sources and reasoning under uncertainty. For example,
\textit{breakout news detection} requires agents to forecast future events;~\textit{resolution detection} requires agents to detect in a noisy world whether a controversy is settled.

\paragraph{Hindsight feedback opportunity.} 
Some tasks in \benchmark{} enable hindsight feedback about an agent's past actions without human intervention. This applies to \textit{breakout news detection} and \textit{popularity prediction}, for which each alerted market's subsequent price path and each recommended post's popularity are visible within hours. 
%In \textit{resolution detection} and \textit{forecasting}, however, feedback is one-shot: each question resolves once, at the end or not at all, and whatever the agent learns it must learn from weaker signals (price drift, analogous questions). 
%Contionus improvement are likely to be helpful on these tasks with recurring feedback. %in contrast, learning components can spend cost budget without benefits in other tasks.

\subsection{Evaluating Agents on \benchmark{}}
Agents interact with each task through a simulated clock that controls which information is available. At time $t$, the environment returns only information that was visible by that point in the replayed stream. The clock advances only when the agent waits, using one of the mechanisms introduced in Sec.~\ref{sec:methods}. Unlike the setup of~\citet{froger2026gaia}, our simulation does not model the LLM's own generation latency. We make this simplification because agents in our tasks act sparsely over simulated periods of several weeks.

At the start of a run, the agent receives an instruction specifying the task, its evaluation metric, and the end of the evaluation window. It must maximize its score over this window while operating under a hard cost budget, and the run terminates if the budget is exhausted. To support budget-aware decisions, the agent receives an updated summary of its spending and remaining budget on each simulated day.
We score the agent deterministically by comparing its timed commitments with the corresponding ground truth. Table~\ref{tab:task-details} in the Appendix provides detailed task descriptions and defines the task-specific evaluation metrics.

%Documented rate limits are enforced as they are in the wild (60 news calls per minute, 1{,}000 social-media calls per ten minutes, each exchange's own limiter including escalating bans), and nothing reports the agent's usage back to it.

%\paragraph{Evaluation Metrics}
%Every task is scored by a deterministic function of the agent's timed commitments $\{\langle y_i, t_i\rangle\}$ and the ground truth fixed at benchmark construction. Table~\ref{tab:task-details} summarizes the metrics.

\section{Experiment Setup}
\label{sec:methods}

Our experiments on \benchmark{} answer the following research questions.
\textbf{RQ1:} How do base LLM and harness designs affect model performance across \benchmark{} tasks? For harness designs, we focus on the mechanism by which agents make timing decisions, and continuous improvement methods.
\textbf{RQ2:} What are notable failure modes of long-lived agents, and how do harness design and continuous improvement address them?

\subsection{Trigger Mechanisms}
% Sec. 4: the two studied harness axes (source: assets/fig_axes/make_fig_axes.py). Requires \usepackage{wrapfig}.
\begin{wrapfigure}{r}{0.5\textwidth}
  \vspace{-\baselineskip}
  \centering
  \includegraphics[width=0.5\textwidth]{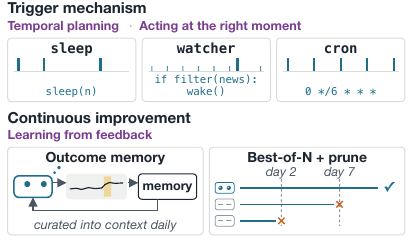}
  \caption{\textbf{Evaluated harness design axes}. Trigger mechanisms affect how agents plan over time and act at the right moment. Continous improvement methods affect how agents use feedback to improve continuously.}
  \label{fig:axes}
  \vspace{-2.5\baselineskip}
\end{wrapfigure}

%Our first evaluation axes is how the agent wakes. The mechanism varies in practice: agents can periodically poll
We vary the trigger mechanisms by providing different tools to the agent for scheduling their next triggers.

\paragraph{LLM-owned sleeping tool (sleep harness).} The ReAct agent is provided a \textit{sleep} tool. The agent can name a duration, and the agent loop continues after the duration expires.

\paragraph{LLM-authored watcher programs (watcher harness).} The agent is provided tools to read, author, and execute a Python watcher program as in the programmatic tool calling or CodeAct~\citep{wang2024executable, Wang2024TroVEIV} paradigm. The harness further allows the program to invoke tools that the agent has access to. The program can exit and hand over control back to the LLM with optional returned results. 

\paragraph{Agent-managed scheduled jobs (cron harness).}  A fixed schedule fires the program, runs a bounded ReAct episode at the current timestamp, and exits when the ReAct episode ends. The agent can list, create, update, and delete its own one-time and recurring triggers through environment endpoints with an optional note describing the intent of the trigger. %We initialize the 

We further note that all three triggering mechanisms are expressive enough to allow the agent to schedule wakes at any timestamp $t \in [0,T]$. Details of the triggering mechanisms and corresponding examples are included in Appendix~\ref{apdx:tm_detail}.

\subsection{Learning from Hindsight Feedback}

We examine a memory-based approach and a parallel-sampling-based approach to learn from hindsight feedback as the agent acts online.

\textbf{Outcome memory.}
Following~\citet{Goel2026FutureSimRW}, after each simulated day, a process curates a memory entry that includes the agent's past commitments, the settled outcomes in the world, and numeric statistics (\textit{e.g.}, the number of false positive alarms), added to the context of the agent. %The memory is inserted to the system prompt after the task instruction. %For breakout news detection, we further examine two variants: (1) self-curated feedback, where the same model labels attributable news given the price breakouts in hindsight. (2) oracle feedback, where a powerful model (GPT-5.6-sol) labels attributable news.
  
%\textbf{Skill curation.}
%Following recent skill curation works that employ propose-validate loops, we implement a skill curation method that proposes skills and optimizes the skill by replaying it over its past trajectory. The skill curation method uses the raw outcome memory as the input. For breakout news detection, we validate the proposed skill in the past $K=40$ wakes, where the agent can decide again whether to commit a breakout news notification given the skill. %For popularity prediction, the agent is evaluated by rolling out on the past simulated day. We perform up to 5 validation loops or until the skill improves over the incumbent skill.

\textbf{Best-of-N + Prune.}~\citep{koh2025tree}
As a simple and cost-efficient approach to test-time scaling with feedback, we simultaneously launch $N=3$ agents at the start. For each day $[t_D,t_{D+1}]$ where $t_D$ is the timestamp of the start of a day, only the commitments of the best-performing agent over $[0,t_D]$ are accepted. We prune agents by performance early on day 2 and day 7 to keep the run under budget. %: from day 2, we stop the agent whose anticipated cumulative spend projects past the budget (unless it is the current best); from day 7, we stop the worse agent among the two. 

Besides memory, we also test periodic harness optimization following recent works~\citep{Lee2026MetaHarnessEO} in the code space. However, we notice that performance improves only with strong optimizer models (\textit{e.g.}, GPT-5.6-sol), which exhaust the budget. We leave the discussion in Appendix~\ref{apdx:online_harness}.

\subsection{Model and Harness Details}
We evaluate eight LLMs under our harness design axes. We only evaluate base LLMs that are cost-efficient when running over simulated weeks (\textit{e.g.,} GPT-5.6-luna, GPT-5.6-terra). We skip more powerful models from the same family (\textit{e.g.,} GPT-5.6-sol) if cheaper models already exhaust $\$20$ in a single run over simulated weeks.

Our experiments control the harness within each \benchmark{} task. For the two tasks requiring a web user interface (\textit{web filing} and \textit{web broker ops}), we use the pinned-version OpenCode harness with Playwright MCP for browser use. For the other tasks, we use a customized agent harness and ReAct agent loop. The agent accumulates context over each wake and discards the earliest interaction history (other than the instruction) when the context exceeds a token limit of $60,000$. We believe more harness axes such as context length can impact cost and performance, and we leave such evaluation as future work.

\section{Experiment Results}

%Seeds affect only in LLM sampling; the replayed world is identical for every run.  %An immedidate observation from Table~\ref{tab:main_table} is the large variance.

%We dissect the variances within runs, and analyze how models and axes can jointly affect mean and variance of the scores.

% Sec. 4.x (source: docs/scripted_analysis/top_methods_bars_v2.py, parsed from results_table_v1.tex).
% Preamble: \usepackage{xcolor} and \usepackage{amssymb} (the harness marks in the caption).
\begin{figure}[t]
  \centering
  \includegraphics[width=\textwidth]{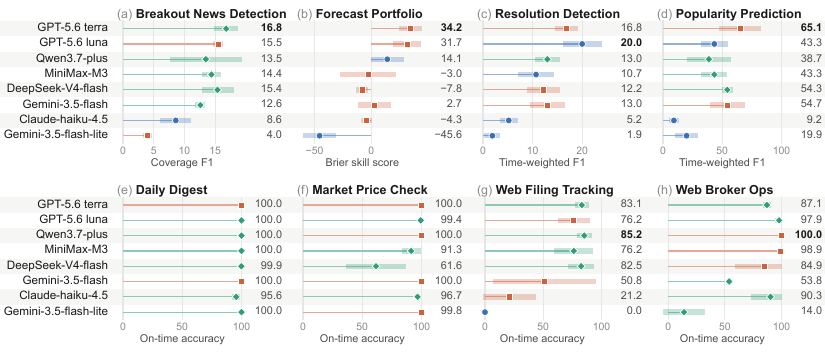}
  \caption{\textbf{Every model under its best triggering mechanism.} Each marker is the harness that scored highest for that model on that task (\textcolor[HTML]{3B6FB6}{$\bullet$}~sleep, \textcolor[HTML]{D1603D}{{\scriptsize$\blacksquare$}}~watcher, \textcolor[HTML]{2F9E77}{$\blacklozenge$}~cron): mean score ($\times 100$) over seeds, band one standard deviation. Full results are in Table~\ref{tab:main_table}.}
  \label{fig:best-harness}
\end{figure}

%Before comparing harness designs and models, we ask how stable agent performance is across repeated runs. Stability is especially important for long-lived agents: compared with consistently mediocre performance, an unexpected failure after weeks of operation can be substantially more costly and surprising to users.
%Table~\ref{tab:main_table} summarizes the results of agents under 8 models and three trigger mechanisms in 3 to 6 repeated runs. %We notice substantial performance variance across models, harness, and even repeated experiments under diffeent seeds.
%Even on routine tasks such as daily digest and market price check, some runs fail catastrophically, yielding high standard deviations (\textit{e.g.}, $81.7 \pm 28.7$ for Claude-Haiku and $89.7 \pm 15.0$ for MiniMax-M3 with the watcher harness). 
%We first analyze the sources of this variation and the effect of model and harness choices.
We start by analyzing the effects of models and trigger mechanisms on agent performance. We then analyze noticeable failure modes of long-lived agents and evaluate how continuous improvement methods address known failure modes of long-lived agents and improve their performance.

%\textbf{Residal variance.} Though we notice two apparent failure modes above, we consider that most of the residual variance belongs to individual agent decisions, rather than to single runs. %Apart from the persistence of poor strategies, the remaining variance arises from the agents deciding differently at each scorable event and is not run-specific. For example, in breakout news detection and resolution detection, seeds of one model and harness are alike in alert volume and spend, yet they anticipate different breakouts or resolution conclusions. 

\subsection{Model and Triggering Mechanism Effects on Performance}
%We identify conclusions that can survive under the variance within the runs.

\textbf{Performance of base LLMs.} Figure~\ref{fig:best-harness} shows the performance of each base LLM under their best triggering mechanism. We notice that the best-performing model is fairly consistent: GPT-5.6-luna or terra performs best on forecasting and detection tasks that require extensive reasoning; Qwen-3.7-plus obtains the best (and even perfect) on-time accuracy in repeated runs on routine tasks such as daily digest, market price check, and web broker ops. 

\begin{figure}[t]
  \centering
  \includegraphics[width=\textwidth]{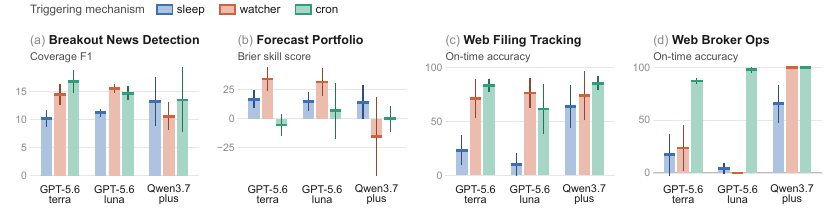}
  \caption{\textbf{Trigger mechanism effect in the models at the frontier performance.} Mean score ($\times 100$) over seeds under each harness (\textcolor[HTML]{3B6FB6}{$\bullet$}~sleep, \textcolor[HTML]{D1603D}{{\scriptsize$\blacksquare$}}~watcher, \textcolor[HTML]{2F9E77}{$\blacklozenge$}~cron), whiskers one standard deviation.}
  \label{fig:trigger-effect-abgh}
\end{figure}

\textbf{Trigger mechanisms affect the performance of agents.}
We analyze the effects of trigger mechanisms on the performance of frontier models. Figure~\ref{fig:trigger-effect-abgh} compares trigger mechanisms across two reasoning-intensive tasks and two routine tasks made more complex by browser use. On (a) breakout news detection, the watcher and cron triggers outperform the sleep trigger by approximately five coverage-F1 points for GPT models, whereas the triggers are indistinguishable for Qwen. On (b) forecast portfolio, the cron trigger performs worse than the sleep trigger. Under cron, agents wake most frequently and repeatedly shift their predictions toward the extremes of yes and no, resulting in many overconfident, incorrect forecasts. On the two browser-use tasks (c) and (d), the sleep trigger can perform catastrophically relative to the cron trigger: agents sometimes skip entire days without performing meaningful work.
Figure~\ref{fig:trigger-stabilize} in the Appendix further visualizes trigger effects on weaker models. On daily digest, Gemini-3.5-flash-lite achieves 100\% accuracy with the cron harness. It also achieves 100\% accuracy with the watcher harness on price check by using programmatic tool calls in the agent-authored program to check prices and issue notifications, while performing poorly under other trigger mechanisms. 
These results show that the trigger mechanism can significantly affect agent performance on both routine and complex tasks, for frontier and weaker models alike.

%When all base LLMs are considered, on breakout news detection, the cron trigger outperforms the watcher trigger in 18 of 24 model$\times$episode cells (paired Wilcoxon $p=0.011$) and the sleep harness in 15 of 24 ($p=0.053$). On forecast portfolio, cron loses to sleep in 7 of 8 models ($p=0.016$) and to watcher in 6 of 8. 

%\textbf{Trigger mechanism can stablize agent performance that otherwise perform catastrophically.} 

%On popularity prediction the harness effect is model-specific: top perfomrance by GPT-5.6-terra (65.1) and Gemini-3.5-flash (54.7) is obtained by the watcher harness, followed by DeepSeek-V4-flash with cron or sleep harness (54.3 and 53.1).
%On resolution detection and popularity detection there is no best triggering mechanism.
%Our results show that on complicated tasks which harness wins is a property of the task$\times$model pair.

% Sec. 4.x (source: docs/scripted_analysis/trigger_effect_v1.py, parsed from results_table_v1.tex).
% Preamble: \usepackage{xcolor} and \usepackage{amssymb} (the harness marks in the caption).

 %later through case studies in Sec.~\ref{ssec:case_study}.

\begin{wrapfigure}{r}{0.5\textwidth}
  \vspace{-\baselineskip}
  \centering
  \includegraphics[width=0.5\textwidth]{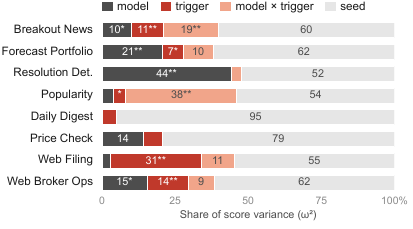}
  \caption{\textbf{Where the score variance goes among the five strongest models per task.} Two-way ANOVA share ($\omega^2$) of per-seed scores attributed to the model, the harness, their interaction, and seed-to-seed variation; $^{*}p<0.05$, $^{**}p<0.01$ (permutation test).}
  \label{fig:anova-tape}
  \vspace{-\baselineskip}
\end{wrapfigure}

\textbf{Variance decomposition of model and trigger mechanism effects.}
We quantitatively evaluate how repeated runs, model choice, trigger mechanism, and their interactions affect agent performance.
Figure~\ref{fig:anova-tape} decomposes per-seed scores into components attributable to the model, trigger, model$\times$trigger interaction, and seed, restricting the analysis to the five strongest models per task.

% Sec. 4.x: variance decomposition, three strongest models per task (source: docs/scripted_analysis/anova_tape_top3_v1.py). Requires \usepackage{wrapfig}.

The decomposition shows that seed-to-seed variation within the same model and trigger configuration is the largest component for every task.
%In repeated runs, the agent faces the same replayed stream, but it can make different decisions.
The differences across repeated runs can be further divided into two parts.
Some differences concern individual scorable events, such as whether to issue an alert; expanding episodes (\textit{e.g.}, time windows) per task can reduce the variance. Other differences affect many decisions over the course of a run. For example, an agent may repeatedly sleep too long, forget its instruction, or exhaust its budget before the episode ends. These persistent behaviors can produce large differences between runs, even when a task contains many scorable events; we consider this source of variance as a challenge of long-lived tasks, rather than evaluation noise. Table~\ref{tab:seed-spread} in the Appendix further reports the overall spread across repeated runs, which differs across trigger mechanisms on several tasks. The case studies in Sec.~\ref{ssec:case_study} further illustrate persistent behaviors within runs.

Among the remaining components, the base LLM dominates for resolution detection (44\%), forecast portfolio (21\%), and web broker ops (15\%).
Models sometimes have their own best trigger mechanism: the contribution of the trigger mechanisms can arise as an interaction effect between the model and the triggering mechanism. 
This model$\times$trigger interaction is significant for breakout news detection (19\%) and popularity prediction (38\%).
On some tasks we also see a trigger main effect, \textit{i.e.,} certain triggers are uniformly better across all base models. This applies to breakout news detection (11\%, with cron ahead of sleep and watcher with Wilcoxon $p=0.039, 0.039$ over 8 models), forecast portfolio (7\%, with cron behind sleep and watcher with $p=0.016, 0.078$), web filing (31\%, cron ahead of sleep with $p=0.031$), and web broker ops (31\%, cron ahead of sleep with $p=0.023$).

\textbf{Triggering mechanisms affect LLM cost, but cost alone does not explain performance.} The harness accounts for 16--28\% of the variance in LLM cost on breakout news detection, resolution detection, and popularity prediction. However, 
across all model and harness configurations, LLM cost and average scores are only weakly correlated ($-0.22 \leq \rho \leq 0.09$). This suggests that the effects of harness design on performance cannot be explained simply by cost.

\subsection{Analysis of Failure Modes}
\label{ssec:case_study}
Our earlier analysis identifies notable failures associated with particular models or trigger mechanisms. We now examine agent execution traces to identify recurring failure modes.

\textbf{Agents can deviate from instructions over long deployments.} On daily digest, an agent may fail after a period of normal operation. In one Claude-Haiku run with the sleep harness, the agent operated normally for the first week, then slept for three weeks and attempted to deliver the remaining outputs on the final day, contrary to the task instructions in the system prompt. Similarly, in a run with the cron harness, the agent independently decided to ``wait until more comments are settled before notification,'' resulting in missed notifications.

\textbf{Strategies chosen early can persist.}
In contrast to late-stage deviations, Reddit popularity prediction exhibits an early, detectable failure mode across models. With the sleep harness, agents often choose their sleep intervals at the beginning of the task and retain them throughout the run; the weekly median sleep interval changes by less than one hour. On Qwen-3.7-plus, a run triggered only every 12 hours performed catastrophically (F1 score${}=5.5$) compared with runs triggered every six hours (F1 score${}=38$). This issue is not limited to the sleep or watcher harnesses. Even under the cron harness, an agent may persist with a poor selection strategy adopted at the outset. For example, one Qwen run with the cron harness consistently performed a no-op at every trigger.

In the next section, we show that test-time feedback can address some of these failures and improve the stability of agent performance.

\subsection{Utilizing Test-Time Feedback}

%\benchmark{} environment provides two types of feedback: daily cost summary (Sec.~\ref{ssec:simu_world}) for all tasks, and hindsight feedback for breakout news detection and popularity prediction. 
%We ask whether test-time feedback can (1) stablize and improve agent's performance, and (2) improve budget usage over time. 

% auto-generated by results_table_v1.ipynb — do not edit by hand
\begin{table}[t]
\definecolor{tablegray}{RGB}{160,160,160}
\definecolor{tablepink}{RGB}{255,35,175}
\newcommand{\tabstd}[1]{\textcolor{tablegray}{_{\pm #1}}}
\newcommand{\tabpct}[1]{\textcolor{tablepink}{(#1\%)}}
\centering
\vspace{-\baselineskip}
\caption{Continous improvement from hindsight feedback: Mean$_{\pm\text{std}}$ over seeds; below each value, mean LLM spend (\$) or, if a seed ran out of budget, the mean share of the window seen (\%). Best method per column shaded.}
\label{tab:results-learning}
\scalebox{0.7}{
\begin{tabular}{llcccccc}
\toprule
Model & Method & \multicolumn{3}{c}{\textbf{Breakout News Detection}} & \multicolumn{3}{c}{\textbf{Popularity Prediction}} \\
 & & \multicolumn{3}{c}{\texttt{coverage F1}} & \multicolumn{3}{c}{\texttt{time-weighted F1}} \\
\cmidrule(lr){3-5} \cmidrule(lr){6-8}
 & & sleep & watcher & cron & sleep & watcher & cron \\
\midrule
GPT-5.6 luna & No learning & \makecell{$11.2\tabstd{0.7}$ \\ \scriptsize (\$4.84)} & \makecell{$15.5\tabstd{0.8}$ \\ \scriptsize (\$7.16)} & \cellcolor{red!18}\makecell{$14.7\tabstd{1.2}$ \\ \scriptsize (\$10.81)} & \makecell{$43.3\tabstd{11.3}$ \\ \scriptsize (\$4.15)} & \makecell{$38.2\tabstd{15.6}$ \\ \scriptsize \tabpct{87}} & \makecell{$41.9\tabstd{5.3}$ \\ \scriptsize (\$7.84)} \\
 & Memory & \cellcolor{red!18}\makecell{$13.0\tabstd{0.7}$ \\ \scriptsize (\$6.82)} & \makecell{$15.5\tabstd{0.8}$ \\ \scriptsize \tabpct{99}} & \makecell{$13.8\tabstd{0.7}$ \\ \scriptsize (\$11.25)} & \makecell{$37.0\tabstd{12.9}$ \\ \scriptsize (\$3.65)} & \cellcolor{red!18}\makecell{$51.1\tabstd{18.9}$ \\ \scriptsize \tabpct{87}} & \cellcolor{red!18}\makecell{$52.1\tabstd{18.0}$ \\ \scriptsize (\$8.39)} \\
 & Best-of-N+Prune & \makecell{$11.6\tabstd{0.8}$ \\ \scriptsize (\$8.04)} & \cellcolor{red!18}\makecell{$16.0\tabstd{0.3}$ \\ \scriptsize (\$9.65)} & \makecell{$14.0\tabstd{0.6}$ \\ \scriptsize (\$16.19)} & \cellcolor{red!18}\makecell{$49.2\tabstd{8.2}$ \\ \scriptsize (\$6.85)} & \makecell{$45.2\tabstd{15.8}$ \\ \scriptsize \tabpct{88}} & \makecell{$42.6\tabstd{1.3}$ \\ \scriptsize (\$14.99)} \\
\midrule
Qwen3.7-plus & No learning & \makecell{$13.2\tabstd{4.4}$ \\ \scriptsize (\$7.56)} & \makecell{$10.6\tabstd{2.5}$ \\ \scriptsize \tabpct{33}} & \cellcolor{red!18}\makecell{$13.5\tabstd{5.8}$ \\ \scriptsize \tabpct{71}} & \makecell{$23.9\tabstd{13.4}$ \\ \scriptsize (\$4.61)} & \makecell{$26.3\tabstd{21.1}$ \\ \scriptsize \tabpct{86}} & \makecell{$38.7\tabstd{18.5}$ \\ \scriptsize (\$10.11)} \\
 & Memory & \makecell{$13.8\tabstd{2.2}$ \\ \scriptsize (\$11.50)} & \cellcolor{red!18}\makecell{$13.6\tabstd{2.4}$ \\ \scriptsize \tabpct{26}} & \makecell{$12.1\tabstd{4.5}$ \\ \scriptsize \tabpct{64}} & \cellcolor{red!18}\makecell{$37.9\tabstd{13.1}$ \\ \scriptsize (\$5.22)} & \makecell{$28.0\tabstd{17.3}$ \\ \scriptsize \tabpct{56}} & \cellcolor{red!18}\makecell{$55.2\tabstd{7.9}$ \\ \scriptsize (\$13.89)} \\
 & Best-of-N+Prune & \cellcolor{red!18}\makecell{$14.4\tabstd{2.9}$ \\ \scriptsize \tabpct{89}} & \makecell{$8.5\tabstd{2.3}$ \\ \scriptsize \tabpct{17}} & \makecell{$13.1\tabstd{2.5}$ \\ \scriptsize \tabpct{42}} & \makecell{$32.1\tabstd{5.4}$ \\ \scriptsize (\$8.42)} & \cellcolor{red!18}\makecell{$37.7\tabstd{15.2}$ \\ \scriptsize \tabpct{82}} & \makecell{$45.7\tabstd{6.7}$ \\ \scriptsize \tabpct{90}} \\
\bottomrule
\end{tabular}}
\end{table}

%We test two mechanisms on the two tasks whose variance has different sources: \emph{memory}, in which the agent reflects daily on the revealed outcomes and carries a learned memory block into later decisions, and \emph{parallel selection}, in which three seeds run concurrently and the seed with the best cumulative hindsight feedback sends the live notifications each day, under one shared budget with dominated seeds pruned or with a private budget per seed (Table~\ref{tab:results-learning}).

%\input{floats/fig-cost-curve}

\textbf{Best-of-$N$ + Prune helps mitigate early failures.}
On popularity prediction, hindsight feedback identifies better policies among seeds within the first few days. Best-of-$N$ selection based on accumulated feedback consistently improves every evaluated cell (\textit{e.g.,} GPT-5.6-luna with sleep, $43.3 \rightarrow 49.2$; Qwen-3.7-plus with sleep, $23.9 \rightarrow 32.1$), halves the variance across seeds under the sleep trigger, and incurs on average $1.61\times$ increase in cost. However, this benefit appears specific to the early-failure mode in popularity prediction: the approach does not consistently improve breakout news detection, which does not exhibit the same failure mode.

\textbf{Effect of memory curated from hindsight feedback.} 
On breakout news detection, memory yields directionally positive changes under the sleep and watcher triggers but a negative change under the cron trigger. On popularity prediction, the change is directionally positive in five of six cells, with GPT-5.6-luna under the sleep harness as the exception. This overall positive tendency suggests that agent memory can improve the performance of long-lived agents. However, its weaker consistency relative to best-of-$N$ sampling indicates that memory alone may be insufficient to redirect an agent after it adopts a poor strategy. Future work can explore cost-efficient methods that help agents use past experience both to accumulate task knowledge and to revise earlier strategies.

We also experiment with continuous harness updates following prior work~\citep{Lee2026MetaHarnessEO,Karten2026ContinualHO}. The algorithm is effective only with strong reflector models, whose cost causes the approach to exhaust its budget early in our setup. We provide further discussion in the Appendix~\ref{apdx:online_harness}.

%As discussed, on popularity prediction, the main failure mode is poor timing decisions chosen early; memories, however, do not help the agent to revise the timing strategy automatically. Breakout news detection does not suffer from the same early-failure mode, yet it requires strong reasoning capabilty to anticipate news and forecast price breaks, which is not improved via a handful of feedback. The ineffectiveness of the hinsight memory contrasts the setups where agents improves by extracting worksflow in tasks such as web navigation~\citep{Ouyang2025ReasoningBankSA,Wang2024AgentWM}. This result implies that hindsight feedback structures in long-lived tasks do not naturally translates into performance improvement.

%\textbf{Agents can adjust costs mid-run.} We further find that injecting cost reports can steer agent behavior more effectively than providing hindsight feedback and can help agents complete tasks within budget. In several repeated runs, the top-performing agent on popularity prediction, GPT-5.6-terra with the watcher harness, responds to a tightening budget by reflecting on its spending and delegating tasks to deterministic watcher programs, as shown in Figure~\ref{fig:cost-curve}.

\section{Related Works}

\paragraph{Long-horizon agents and temporal awareness.}
Agent benchmarks have expanded from bounded software and tool-use tasks~\citep{Jimenez2023SWEbenchCL, Yao2024benchAB} to extended coding and research tasks~\citep{Yang2026ProgramBenchCL, Zhu2026EdgeBenchUS}. A complementary line studies agents in environments that change over time: TimeArena and Robotouille model action durations and asynchronous execution~\citep{Zhang2024TimeArenaSE, GonzalezPumariega2025RobotouilleAA}, while real-time reasoning evaluates the tradeoff between reasoning depth and response latency~\citep{Wen2025RealTimeRA}. GAIA2 evaluates dynamic, asynchronous interactions~\citep{froger2026gaia}, and SentinelBench directly compares waiting mechanisms for monitoring tasks using success, reaction time, and resource use~\citep{Maldaner2026SentinelBenchAB}. Closely related work studies lightweight triggers for LLM invocation~\citep{Liu2026DoPA} and the execution of deferred intentions across a simulated week~\citep{Liu2026PMBenchEP}. Together with evidence of temporal misalignment in tool use~\citep{Cheng2025YourLA}, these works motivate evaluating timing alongside task completion. \benchmark{} extends this evaluation to recurring obligations and sparse actions over weeks of replayed real-world streams under a cumulative budget. Related problems also arise in stream monitoring~\citep{Lin2016TrecRTS}, event- and self-triggered control~\citep{Heemels2012TriggeredControl}, and change detection with costly observations~\citep{Banerjee2012ObservationControl}.

\paragraph{Agent evaluation on live and replayed worlds.}
Forecasting benchmarks evaluate predictions against subsequently resolved events~\citep{Zeng2025FutureXAA, Bianchi2025Back, karger2025forecastbench, Yang2025LLMasaProphetUP}, while financial agents are evaluated on live markets and historical data~\citep{Qian2025WhenAT, yu2024fincon, Li2024INVESTORBENCHAB, Chen2025StockBenchCL}.
~\citet{Chai2026AgenticTM} simulates the state of the web at any specific past date through post-cutoff agentic content filtering.
Particularly close to our setting, FutureSim replays news chronologically over 90 days and lets agents revise forecasts and adapt from resolved outcomes; its reported environment advances at a daily cadence~\citep{Goel2026FutureSimRW}. \benchmark{} shares the use of replay, while evaluating finer-grained trigger decisions across forecasting, detection, and routine automation tasks.

\paragraph{Test-time continuous improvement.}
Agents can improve through reflection on execution feedback~\citep{Shinn2023ReflexionLA, Zhao2023ExpeLLA} and reuse experience through memories, workflows, or skills~\citep{Wang2024AgentWM, Ouyang2025ReasoningBankSA, Zheng2025SkillWeaverWA}. Live-Evo studies memory evolution from continuous forecasting feedback~\citep{Zhang2026LiveEvoOE}, and EvoArena evaluates memory evolution in dynamic environments~\citep{Xu2026EvoArenaTM}. Other methods optimize prompts~\citep{Agrawal2025GEPARP}, skills~\citep{Yang2026SkillOptES}, harnesses~\citep{Lee2026MetaHarnessEO, Wei2026EvoHarnessCS}, or entire agent programs~\citep{Zhang2025DarwinGM, Yin2025GdelAA}; Adaptive Auto-Harness extends harness adaptation to open-ended task streams~\citep{Liu2026AdaptiveAS}. Complementary work scales agent inference through search and multiple candidate trajectories~\citep{koh2025tree, GonzalezPumariega2025ScalingAF}. %Our experiments examine outcome memory and online selection among parallel trajectories using hindsight feedback. We evaluate their benefits together with their inference costs within the same budget used to carry out the long-lived task.

%\paragraph{Connections to earlier monitoring research.}
 %These provide background on timing and observation tradeoffs; our empirical focus is the behavior of contemporary LLM agents and their harnesses on long-lived tasks.

\section{Conclusion}

We introduced \benchmark{}, an evaluation environment that places agents in a chronologically replayed world for weeks. We separated two harness decisions: when the agent wakes and how it uses hindsight feedback, and evaluated three trigger mechanisms and two continuous improvement methods across eight LLMs with repeated runs per configuration. 
Our experiments demonstrate that base models and trigger mechanisms have an impact on the performance of agents via standalone effects or their interactions.
Best-of-$N$ selection with pruning mitigates the early failures that reflection leaves in place, while memories curated from hindsight feedback improve performance less consistently. We hope our paper motivates future research on more stable, performant, and cost-effective models and harnesses for long-lived agents deployed in real-world, online, dynamic environments.

%Repeated runs of the same configuration vary substantially, and with three to six seeds per cell this variance bounds what \benchmark{} can resolve. The variance has two sources. On the news-related tasks it is the expected spread of independent hits and misses over a limited number of scored events; on popularity prediction and daily digest it comes from a few runs that commit to a poor strategy early and hold it for weeks. The next release will add scored events per task, spread across the episode and with partial credit for near-miss timing, which reduces the first source, and will run six seeds per cell by default, which reduces the second. We also evaluated only three trigger mechanisms and two learning mechanisms; extending the catalogue to further test-time improvement methods is left for future work.

\textbf{Limitations.} This paper is an evaluation study rather than a leaderboard. The replay environment and data pipeline support arbitrarily many models, harnesses, and repeated runs; we report three to six runs per configuration, run 1 to 3 episodes per task, and draw conclusions based on comparisons aggregated across models or harness design axes. 
We do not rank individual model-and-trigger pairs in this work.
%The large spread across repeated runs of the same configuration is a finding about long-lived agents, with noise from the small number of seeds and evaluated episodes (time spans or markets). 
Future work that needs to perform benchmarking and separate methods at the level of individual cells can run more seeds per configuration and replay additional episodes to add scorable events; the environment supports both without any change to the tasks or metrics.

In addition, we evaluated only two axes of harness design: triggering mechanisms and test-time learning methods. Evaluation along additional axes, such as context management, would be valuable. %We will also study the ability of LLMs to design task- and model-specific harnesses freely in program space, rather than selecting among predefined options along fixed axes. 
Furthermore, within these two axes of harness design, we evaluated only three triggering mechanisms and two test-time learning approaches. Expanding the set of methods is left for future work.

\section*{AI Use Statement}
In this work, we used AI assistance to implement the experimental framework and methods. Human authors used AI coding agents to implement an agent-interactive environment with simulated time, pipelines for batch processing raw data (such as Common Crawl news), the agent loop used by task-solving agents, triggering mechanisms, and continuous-improvement algorithms. All code was reviewed by the human authors.

The experiments involve three non-code artifacts created with AI assistance: (1) the initial task instructions provided to the evaluated agents and (2) the synthetic trades placed against replayed real-world market data in the \textit{web broker ops} task, (3) the web portals that \textit{web filing tracking} and \textit{web broker ops} task operates on. All artifacts were reviewed and revised by the human authors.

During the early stages of the project, we also used AI assistance for brainstorming, defining the scope of the paper, and identifying related work. Later, AI assisted with statistical and qualitative analyses of raw experimental records, including agents' final scores and action trajectories. AI also assisted in polishing writing and generating vector graphics for this paper.

\section*{Reproducibility statement}

We release code and data curation scripts. All required data is publicly accessible and permissively licensed; details in Appendix~\ref{apdx:license}.

\bibliography{iclr2027_conference}
\bibliographystyle{iclr2027_conference}

\appendix
\newpage

% Task reference, one row per task (name + sketch | description + metric); breaks across pages.
% Task order and task/metric names follow the results table (tab:main_table): bnpm, fp, rd, reddit,
% digest, cpc, edgar, broker.
% Requires: xltabular (loads longtable + tabularx), booktabs, array, graphicx, xcolor.
% Not a float: it starts where it is \input (put a \clearpage before it in the appendix). One row per
% task so each row's height includes its sketch and page breaks fall only between tasks.
% Sketches in column 1 are synthetic (docs/paper_draft/assets/task_description/make_task_sketches.py).
% scoreshade = the figures' green fill (#2a9d4f @ 30%) flattened on white.
\definecolor{scoreshade}{RGB}{191,226,202}
\begingroup
\setlength{\tabcolsep}{5pt}
\renewcommand{\arraystretch}{1.15}
\setlength{\LTcapwidth}{\textwidth}
\begin{xltabular}{\textwidth}{@{} >{\small\raggedright\arraybackslash}p{3.3cm} >{\small\raggedright\arraybackslash}X @{}}
\caption{The eight \benchmark{} tasks: what the agent must do (world, commitment, horizon) and the primary metric. The schematic sketches show the signal the agent watches over time, with the \colorbox{scoreshade}{shade} (or dots) marking when a commitment earns credit, fading as the credit decays over time.}
\label{tab:task-details}\\
\toprule
Task & Description and primary metric \\
\midrule
\endfirsthead
\caption[]{(continued)}\\
\toprule
Task & Description and primary metric \\
\midrule
\endhead
\multicolumn{2}{r@{}}{\footnotesize\itshape continued on next page}\\
\endfoot
\bottomrule
\endlastfoot

% ---------------------------------------------------------------- bnpm
\textbf{Breakout News Detection}\par\vspace{4pt}
  \includegraphics[width=\linewidth]{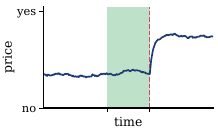}
& \textbf{Task description.}
The agent monitors 8 Polymarket markets over a four-week episode, with the BM25 news search API and each market's minute-resolution price history up to the current time. A \emph{breakout} is an abrupt price move following the definition of~\citet{Yu2026BuildingSW}: a day-over-day change that is statistically large against the market's recent volatility, localized to the minute it starts. A commitment is a notification $\langle \text{market}, \text{article}, \text{direction}\rangle$ claiming that the market will start a breakout in that direction within the next 24\,h, citing the news article as evidence. A notification stands until a breakout in the named market starts or 24\,h pass, and only one notification per market may stand at a time. A notification placed at or after the move starts earns nothing. The task is run over three episodes with disjoint markets and days, March 1--28, March 28--April 24 and April 24--May 21, 2026, holding 25, 30 and 33 ground-truth breakpoints (88 in total); scores pool the three episodes per seed.
\par\smallskip
\textbf{Metric: coverage F1.}
A breakpoint is \emph{covered} if a notification with the right market and direction was standing in the 24\,h \textit{before} it started. Recall $=$ covered breakpoints / all breakpoints; precision $=$ covering notifications / all resolved notifications. \\
\midrule

% ---------------------------------------------------------------- fp
\textbf{Forecast Portfolio}\par\vspace{4pt}
  \includegraphics[width=\linewidth]{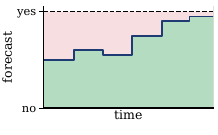}
& \textbf{Task description.}
The agent maintains standing probability forecasts on a portfolio of 60 real-world Polymarket questions (elections, conflicts, economic events) over 31 days (March 2026), with the news API; market prices are \textbf{not provided}. Questions enter the world at their real open date and may appear at any time. A commitment is a probability distribution over \texttt{yes} or \texttt{no}. Setup  adapted from FutureSim~\cite{Goel2026FutureSimRW}.
\par\smallskip
\textbf{Metric: Brier skill score.}
At every instant of a question's life its standing forecast $p$ incurs the Brier score $\sum_o (p_o - \mathbf{1}[o\ \text{resolved}])^2$; the per-question score is the exact time integral of this step function over the question's visible window divided by its length, and the run's Brier score $\mathrm{BS}$ is the mean over questions that resolved in-window (35 here). The skill score is $1-\mathrm{BS}/\mathrm{BS}_{\text{base}}$, where $\mathrm{BS}_{\text{base}}$ is the Brier score of always forecasting the base rate of \texttt{yes} among the resolved questions (0.2 here). A perfect forecast scores 1, the base-rate forecast 0, and the uniform forecast over \texttt{yes} and \texttt{no} $-0.56$. \\
\midrule

% ---------------------------------------------------------------- rd
\textbf{Resolution Detection}\par\vspace{4pt}
  \includegraphics[width=\linewidth]{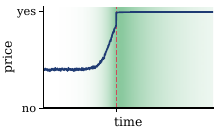}
& \textbf{Task description.}
The agent watches a portfolio of 60 standing Polymarket questions (elections, conflicts, corporate and economic events) over a 45-day episode with the news API. Market prices and resolutions are \textbf{not provided}: the agent must judge from news alone whether the world has already settled a question. Questions enter the portfolio at their real open date. In each episode 27 questions resolve inside the window and 33 do not; the latter stay open throughout, so any claim on them is a false alarm. A commitment is a claim, with the outcome, that a question is \emph{already decided} in the world, as opposed to a forecast of how it will resolve. The agent gets one claim per question, ever; it cannot be revised or withdrawn, and questions decided during the run but never claimed count as misses. The task is run over three episodes with disjoint questions, March 1--April 15, April 10--May 25 and May 16--June 30, 2026 (81 in-window resolutions in total); scores pool the three episodes per seed.
\par\smallskip
\textbf{Metric: time-weighted F1.}
Ground truth is the earliest instant $t^\ast$ at which the outcome was decided according to Polymarket (with a grace window). A correct claim at time $t_i$ earns credit $e^{-(t^\ast - t_i)/12\text{h}}$ if early and $e^{-(t_i - t^\ast)/48\text{h}}$ if late. %(\textit{i.e.,} anticipating a resolution is penalised four times harder per hour than reporting it late).
Recall $=$ total credit / decided questions, precision $=$ total credit / claims issued. \\
\midrule

% ---------------------------------------------------------------- reddit
\textbf{Popularity Prediction}\par\vspace{4pt}
  \includegraphics[width=\linewidth]{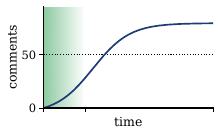}
& \textbf{Task description.}
The agent watches a live replay of ten AI-focused subreddits for 30 days (April 2026); new posts arrive continuously and each post's comment cascade is served as its causal prefix. A commitment is a recommendation of one post predicted to become \emph{popular}, defined as $\ge 50$ comments 24\,h after posting; at most $K=10$ recommendations are accepted per rolling 24\,h.
\par\smallskip
\textbf{Metric: time-weighted F1.}
Each popular post that was recommended earns weight $\max(0,\,1-\text{latency}/6\text{h})$ for its first recommendation, where latency is measured from the post's creation---full credit at posting, none after six hours. Recall $=$ total weight / the cap-limited oracle total ($\sum_{\text{days}} \min(K, \#\text{popular posts that day})$); precision $=$ popular recommendations / all settled recommendations. \\
\midrule

% ---------------------------------------------------------------- digest
\textbf{Daily Digest}\par\vspace{4pt}
  \includegraphics[width=\linewidth]{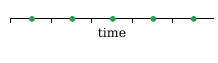}
& \textbf{Task description.}
Same subreddit replay and 30-day window (April 2026). Once a day the agent delivers one digest of 10 popular ($\ge 50$ comments) posts from yesterday.
\par\smallskip
\textbf{Metric: on-time accuracy.}
For each scored day, the first 10 distinct ids of the accepted digest are checked: the day's score is the fraction of them that lie in the candidate window and are popular. A day without a digest scores 0. \\
\midrule

% ---------------------------------------------------------------- cpc
\textbf{Market Price Check}\par\vspace{4pt}
  \includegraphics[width=\linewidth]{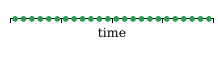}
& \textbf{Task description.}
Every hour for 14 days (March 1--15, 2026) the agent must report the current spot price of three given assets. Prices come from three exchanges' public 5-minute candle endpoints, replayed through a proxy that returns raw bodies and injects an incident stream fitted to public exchange status-page archives.
The window holds 9 provider incidents (5 maintenance windows of 14--201\,min, one 30\,min outage, three degraded stretches of 47--318\,min on one venue).
\par\smallskip
\textbf{Metric: on-time accuracy.}
The reference price for a report is the cross-venue median at the moment of submission. A symbol-hour scores 1 if the report is within 5 basis points of the reference and 0 otherwise.
%The run score is the mean over the before the run's LLM budget was exhausted.
\\
\midrule

% ---------------------------------------------------------------- edgar
\textbf{Web Filing Tracking}\par\vspace{4pt}
  \includegraphics[width=\linewidth]{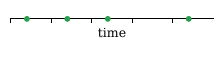}
& \textbf{Task description.}
The agent keeps a fund's internal portfolio web app current for 33 holdings over April 1--June 1, 2026, reading a replay of SEC EDGAR's JSON endpoints served as of the current time. A commitment is a filing record entered through the app's web form in a browser: for each held company's earnings press release (8-K, item 2.02) and periodic report (10-Q/10-K), the accession number, form and acceptance time, plus, for periodic reports, the fiscal period, revenue and diluted EPS from the filing's XBRL facts. 63 filings count in this window. Following WebArena~\citep{zhou2024webarena}, we created synthetic web form app.
\par\smallskip
\textbf{Metric: on-time accuracy.}
Each filing is checked once, at the first NYSE open after its acceptance. It scores 1 if its record exists, matches the filing's XBRL values within tolerance, and was saved after the filing was accepted and after the agent fetched that company's data from the replay; otherwise 0. The score is the mean over filings. \\
\midrule

% ---------------------------------------------------------------- broker
\textbf{Web Broker Ops}\par\vspace{4pt}
  \includegraphics[width=\linewidth]{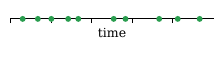}
& \textbf{Task description.}
The agent keeps a levered crypto account (BTC, ETH; replayed 5-minute exchange candles) in good standing on a broker web portal for 14 days (April 1--15, 2026). The manager's book trades on a fixed schedule of day-orders; the agent does not trade. All actions go through the portal's forms in a browser. A commitment is a duty done in time: an opposite-side protective stop within 2\,h of each fill, and a funding instruction within 4\,h of each margin call (issued when the margin ratio falls below the broker's requirement, which the broker raises twice by notice). Every trade event is announced via synthetic event API accessible by the agent. The window holds 31 duty instances (29 fills, 2 margin calls) at instants spread across the day.
\par\smallskip
\textbf{Metric: on-time accuracy.}
Each duty instance is checked once, at its deadline, and scores 1 if it was met. The score is met instances / all instances. \\
\end{xltabular}
\endgroup

\section{Triggering Mechanism Details}
\label{apdx:tm_detail}

\textbf{Sleep harness.} In the sleep harness, the agent is provided with a \texttt{sleep} tool. The environment fast-forwards its time to the specified time when the agent invokes the tool. Table~\ref{tab:wait-tma-luna} exemplifies an agent trace in the first simulated week on the breakout news detection task. 

\textbf{Watcher harness.} Table~\ref{tab:wait-tmb-luna} illustrates an agent trace with the watcher harness. In this harness, the agent authors a watcher program in Python and is provided with the \texttt{list\_dir}, \texttt{read\_file}, \texttt{write\_file}, \texttt{edit\_file}, and \texttt{run\_program} tools. The agent can access its workspace directory, which is initialized with \texttt{envkit}, a library that allows agent-authored Python programs to invoke the same tools available to the agent. The workspace also includes an agent skill that instructs LLMs on how to write watcher programs using \texttt{envkit}. The skill comprises a Markdown instruction file and two example scripts: \texttt{sleep.py}, which sleeps until a specified timestamp, and \texttt{gatekeeper.py}, a basic task-specific condition watcher with placeholder entity names.
Figure~\ref{lst:example-gatekeeper} shows an example of the default watcher program, and Figure~\ref{lst:fed-watch-luna} shows a program authored by GPT-5.6-luna.

In a ReAct loop, the agent authors the watcher program and invokes the \texttt{run\_program} tool to execute the program. This tool call blocks until the program exits, or until \texttt{envkit.handover} (L22, Figure~\ref{lst:fed-watch-luna}) is called in the program, which adds its argument to the tool call output. The simulated time of the environment passes only via \texttt{envkit.wait} called in the program (L23, Figure~\ref{lst:fed-watch-luna}). A loop that never waits is terminated after $200,000$ executed lines of the agent-authored code, with library calls counting as a single line.

\textbf{Cron harness.} In the cron harness, the agent is triggered according to a cron schedule. The schedule table is initialized with a task-specific baseline schedule whose minimum trigger frequency is determined by the authors (\textit{e.g.,} every six hours for breakout news detection and daily for the daily digest task). This baseline schedule is read-only for the agent, but the agent can create and update additional recurring or one-time trigger schedules. Table~\ref{tab:wait-tmd-luna} illustrates an agent trace.

% Sec. 4.x (source: docs/scripted_analysis/trigger_stabilize_v1.py, parsed from results_table_v1.tex).
\begin{figure}[t]
  \centering
  \includegraphics[width=\textwidth]{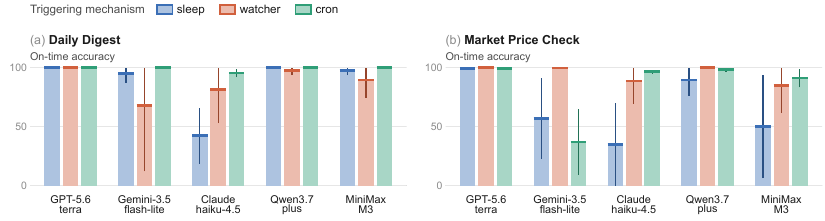}
  \caption{\textbf{The trigger lifts and steadies models below the frontier.} Mean score ($\times 100$) over seeds under each triggering mechanism, whiskers one standard deviation. GPT-5.6 terra is at the ceiling under any trigger; the other models reach it only under the right one.}
  \label{fig:trigger-stabilize}
\end{figure}

% Appendix: LLM cost rates used to book every run's spend (source: configs/model_costs.yaml).
\begin{table}[t]
\centering
\small
\setlength{\tabcolsep}{6pt}
\caption{LLM cost rates in \$ per million tokens. Every run's LLM spend, including the budget it is charged against, is computed from these fixed rates; cached input is the prompt-cache read rate.}
\label{tab:cost-rates}
\begin{tabular}{@{}llrrr@{}}
\toprule
Model & Provider & Input & Cached input & Output \\
\midrule
GPT-5.6 luna       & OpenAI    & 0.20 & 0.02  & 1.20  \\
GPT-5.6 terra      & OpenAI    & 2.00 & 0.20  & 12.00 \\
%GPT-5.6 sol        & OpenAI    & 5.00 & 0.50  & 30.00 \\
Gemini-3.5-flash-lite & Google & 0.30 & 0.03  & 2.50  \\
Gemini-3.5-flash   & Google    & 1.50 & 0.15  & 9.00  \\
Claude-Haiku-4.5   & Anthropic & 1.00 & 0.10  & 5.00  \\
Qwen-3.7-plus      & Alibaba   & 0.32 & 0.064 & 1.28  \\
MiniMax-M3         & MiniMax   & 0.30 & 0.06  & 1.20  \\
DeepSeek-V4-flash  & DeepSeek  & 0.22 & 0.022 & 0.66  \\
\bottomrule
\end{tabular}
\end{table}

% auto-generated by results_table_v1.ipynb — do not edit by hand
\begin{table}[t]
\definecolor{tablegray}{RGB}{160,160,160}
\definecolor{tablepink}{RGB}{255,35,175}
\newcommand{\tabstd}[1]{\textcolor{tablegray}{_{\pm #1}}}
\newcommand{\tabpct}[1]{\textcolor{tablepink}{(#1\%)}}
\centering
\caption{Primary metric per model and timing mode (mean$_{\pm\text{std}}$ over 3 to 6 seeds). Below each value: (\$) mean LLM spend when no seed ran out of budget, or (\%) otherwise: the mean share of the simulated window survived across all seeds.}
\label{tab:main_table}
\scalebox{0.55}{
\begin{tabular}{lcccccccccccc}
\toprule
Model & \multicolumn{3}{c}{\textbf{Breakout News Detection}} & \multicolumn{3}{c}{\textbf{Forecast Portfolio}} & \multicolumn{3}{c}{\textbf{Resolution Detection}} & \multicolumn{3}{c}{\textbf{Popularity Prediction}} \\
 & \multicolumn{3}{c}{\texttt{coverage F1}} & \multicolumn{3}{c}{\texttt{brier skill score}} & \multicolumn{3}{c}{\texttt{time-weighted F1}} & \multicolumn{3}{c}{\texttt{time-weighted F1}} \\
\cmidrule(lr){2-4} \cmidrule(lr){5-7} \cmidrule(lr){8-10} \cmidrule(lr){11-13}
 & sleep & watcher & cron & sleep & watcher & cron & sleep & watcher & cron & sleep & watcher & cron \\
\midrule
GPT-5.6 luna & \makecell{$11.2\tabstd{0.7}$ \\ \scriptsize (\$4.84)} & \cellcolor{red!24}\makecell{$15.5\tabstd{0.8}$ \\ \scriptsize (\$7.16)} & \cellcolor{red!6}\makecell{$14.7\tabstd{1.2}$ \\ \scriptsize (\$10.81)} & \cellcolor{red!12}\makecell{$15.0\tabstd{8.2}$ \\ \scriptsize (\$0.26)} & \cellcolor{red!24}\makecell{$31.7\tabstd{12.2}$ \\ \scriptsize (\$0.20)} & \makecell{$7.0\tabstd{24.1}$ \\ \scriptsize (\$0.41)} & \cellcolor{red!30}\makecell{$20.0\tabstd{3.9}$ \\ \scriptsize (\$1.20)} & \cellcolor{red!24}\makecell{$17.6\tabstd{5.8}$ \\ \scriptsize (\$2.78)} & \cellcolor{red!18}\makecell{$17.1\tabstd{4.6}$ \\ \scriptsize (\$3.27)} & \makecell{$43.3\tabstd{11.3}$ \\ \scriptsize (\$4.15)} & \makecell{$38.2\tabstd{15.6}$ \\ \scriptsize \tabpct{87}} & \makecell{$41.9\tabstd{5.3}$ \\ \scriptsize (\$7.84)} \\
GPT-5.6 terra & \makecell{$10.2\tabstd{1.5}$ \\ \scriptsize \tabpct{66}} & \makecell{$14.5\tabstd{1.9}$ \\ \scriptsize \tabpct{59}} & \cellcolor{red!30}\makecell{$16.8\tabstd{2.0}$ \\ \scriptsize \tabpct{41}} & \cellcolor{red!18}\makecell{$16.6\tabstd{7.8}$ \\ \scriptsize (\$4.23)} & \cellcolor{red!30}\makecell{$34.2\tabstd{10.0}$ \\ \scriptsize \tabpct{91}} & \makecell{$-5.4\tabstd{9.8}$ \\ \scriptsize (\$6.79)} & \makecell{$11.5\tabstd{3.4}$ \\ \scriptsize (\$11.41)} & \cellcolor{red!12}\makecell{$16.8\tabstd{2.3}$ \\ \scriptsize \tabpct{93}} & \cellcolor{red!6}\makecell{$15.3\tabstd{6.8}$ \\ \scriptsize \tabpct{89}} & \makecell{$37.6\tabstd{13.6}$ \\ \scriptsize \tabpct{69}} & \cellcolor{red!30}\makecell{$65.1\tabstd{17.9}$ \\ \scriptsize \tabpct{90}} & \makecell{$18.8\tabstd{7.2}$ \\ \scriptsize \tabpct{25}} \\
Gemini-3.5-flash-lite & \makecell{$1.2\tabstd{2.2}$ \\ \scriptsize (\$0.86)} & \makecell{$4.0\tabstd{0.8}$ \\ \scriptsize (\$9.58)} & \makecell{$3.6\tabstd{0.1}$ \\ \scriptsize (\$14.28)} & \makecell{$-45.6\tabstd{14.6}$ \\ \scriptsize (\$0.11)} & \makecell{$-62.0\tabstd{5.6}$ \\ \scriptsize (\$0.70)} & \makecell{$-58.8\tabstd{4.4}$ \\ \scriptsize (\$0.20)} & \makecell{$1.9\tabstd{1.6}$ \\ \scriptsize (\$0.14)} & \makecell{$0.1\tabstd{0.1}$ \\ \scriptsize \tabpct{94}} & \makecell{$0.0\tabstd{0.0}$ \\ \scriptsize (\$0.08)} & \makecell{$19.9\tabstd{9.6}$ \\ \scriptsize (\$2.39)} & \makecell{$18.8\tabstd{0.7}$ \\ \scriptsize \tabpct{45}} & \makecell{$1.2\tabstd{1.2}$ \\ \scriptsize (\$7.20)} \\
Gemini-3.5-flash & \makecell{$10.4\tabstd{3.6}$ \\ \scriptsize \tabpct{38}} & \makecell{$7.1\tabstd{1.0}$ \\ \scriptsize \tabpct{8}} & \makecell{$12.6\tabstd{0.8}$ \\ \scriptsize \tabpct{27}} & \makecell{$-20.8\tabstd{8.5}$ \\ \scriptsize \tabpct{88}} & \makecell{$2.7\tabstd{14.2}$ \\ \scriptsize \tabpct{34}} & \makecell{$-50.4\tabstd{8.4}$ \\ \scriptsize \tabpct{97}} & \makecell{$7.3\tabstd{3.6}$ \\ \scriptsize (\$6.06)} & \makecell{$13.0\tabstd{3.6}$ \\ \scriptsize \tabpct{90}} & \makecell{$7.0\tabstd{1.6}$ \\ \scriptsize (\$7.36)} & \makecell{$35.4\tabstd{4.8}$ \\ \scriptsize \tabpct{74}} & \cellcolor{red!24}\makecell{$54.7\tabstd{14.7}$ \\ \scriptsize \tabpct{91}} & \makecell{$16.9\tabstd{1.0}$ \\ \scriptsize \tabpct{18}} \\
Claude-haiku-4.5 & \makecell{$8.6\tabstd{2.5}$ \\ \scriptsize \tabpct{62}} & \makecell{$5.2\tabstd{2.3}$ \\ \scriptsize \tabpct{14}} & \makecell{$6.8\tabstd{0.9}$ \\ \scriptsize \tabpct{10}} & \makecell{$-14.1\tabstd{19.1}$ \\ \scriptsize (\$12.18)} & \makecell{$-4.3\tabstd{4.8}$ \\ \scriptsize \tabpct{99}} & \makecell{$-9.4\tabstd{10.3}$ \\ \scriptsize \tabpct{78}} & \makecell{$5.2\tabstd{1.8}$ \\ \scriptsize (\$6.23)} & \makecell{$4.7\tabstd{1.8}$ \\ \scriptsize \tabpct{95}} & \makecell{$2.6\tabstd{0.6}$ \\ \scriptsize \tabpct{99}} & \makecell{$9.2\tabstd{4.5}$ \\ \scriptsize \tabpct{79}} & \makecell{$6.4\tabstd{3.6}$ \\ \scriptsize \tabpct{78}} & \makecell{$4.2\tabstd{3.0}$ \\ \scriptsize \tabpct{23}} \\
Qwen3.7-plus & \makecell{$13.2\tabstd{4.4}$ \\ \scriptsize (\$7.56)} & \makecell{$10.6\tabstd{2.5}$ \\ \scriptsize \tabpct{33}} & \makecell{$13.5\tabstd{5.8}$ \\ \scriptsize \tabpct{71}} & \cellcolor{red!6}\makecell{$14.1\tabstd{14.6}$ \\ \scriptsize (\$1.02)} & \makecell{$-15.1\tabstd{33.9}$ \\ \scriptsize (\$1.34)} & \makecell{$-0.1\tabstd{11.4}$ \\ \scriptsize (\$1.07)} & \makecell{$3.9\tabstd{1.4}$ \\ \scriptsize (\$0.96)} & \makecell{$7.1\tabstd{6.2}$ \\ \scriptsize \tabpct{83}} & \makecell{$13.0\tabstd{2.5}$ \\ \scriptsize (\$2.21)} & \makecell{$23.9\tabstd{13.4}$ \\ \scriptsize (\$4.61)} & \makecell{$26.3\tabstd{21.1}$ \\ \scriptsize \tabpct{86}} & \makecell{$38.7\tabstd{18.5}$ \\ \scriptsize (\$10.11)} \\
MiniMax-M3 & \makecell{$12.1\tabstd{2.5}$ \\ \scriptsize (\$7.83)} & \makecell{$4.4\tabstd{3.8}$ \\ \scriptsize \tabpct{74}} & \makecell{$14.4\tabstd{1.6}$ \\ \scriptsize \tabpct{79}} & \makecell{$-9.2\tabstd{20.7}$ \\ \scriptsize (\$2.30)} & \makecell{$-3.0\tabstd{24.5}$ \\ \scriptsize (\$3.32)} & \makecell{$-14.8\tabstd{18.1}$ \\ \scriptsize (\$2.31)} & \makecell{$10.7\tabstd{3.7}$ \\ \scriptsize (\$3.49)} & \makecell{$6.4\tabstd{3.0}$ \\ \scriptsize (\$7.68)} & \makecell{$9.7\tabstd{1.6}$ \\ \scriptsize (\$2.57)} & \makecell{$41.3\tabstd{19.1}$ \\ \scriptsize (\$5.90)} & \makecell{$26.8\tabstd{3.5}$ \\ \scriptsize \tabpct{92}} & \cellcolor{red!6}\makecell{$43.3\tabstd{10.9}$ \\ \scriptsize \tabpct{99.9}} \\
DeepSeek-V4-flash & \cellcolor{red!12}\makecell{$15.1\tabstd{2.5}$ \\ \scriptsize (\$7.32)} & \makecell{$13.5\tabstd{2.1}$ \\ \scriptsize \tabpct{88}} & \cellcolor{red!18}\makecell{$15.4\tabstd{2.6}$ \\ \scriptsize (\$16.59)} & \makecell{$-11.3\tabstd{17.5}$ \\ \scriptsize (\$1.65)} & \makecell{$-7.8\tabstd{5.3}$ \\ \scriptsize (\$3.14)} & \makecell{$-26.7\tabstd{6.9}$ \\ \scriptsize (\$1.28)} & \makecell{$10.7\tabstd{2.0}$ \\ \scriptsize (\$1.33)} & \makecell{$12.2\tabstd{3.3}$ \\ \scriptsize (\$3.59)} & \makecell{$11.7\tabstd{1.4}$ \\ \scriptsize (\$1.15)} & \cellcolor{red!12}\makecell{$53.1\tabstd{4.3}$ \\ \scriptsize (\$5.45)} & \makecell{$36.3\tabstd{22.0}$ \\ \scriptsize (\$4.31)} & \cellcolor{red!18}\makecell{$54.3\tabstd{4.9}$ \\ \scriptsize (\$12.24)} \\
\midrule
Model & \multicolumn{3}{c}{\textbf{Daily Digest}} & \multicolumn{3}{c}{\textbf{Market Price Check}} & \multicolumn{3}{c}{\textbf{Web Filing Tracking}} & \multicolumn{3}{c}{\textbf{Web Broker Ops}} \\
 & \multicolumn{3}{c}{\texttt{on-time accuracy}} & \multicolumn{3}{c}{\texttt{on-time accuracy}} & \multicolumn{3}{c}{\texttt{on-time accuracy}} & \multicolumn{3}{c}{\texttt{on-time accuracy}} \\
\cmidrule(lr){2-4} \cmidrule(lr){5-7} \cmidrule(lr){8-10} \cmidrule(lr){11-13}
 & sleep & watcher & cron & sleep & watcher & cron & sleep & watcher & cron & sleep & watcher & cron \\
\midrule
GPT-5.6 luna & \makecell{$98.9\tabstd{2.0}$ \\ \scriptsize (\$0.43)} & \makecell{$98.9\tabstd{2.0}$ \\ \scriptsize (\$0.16)} & \cellcolor{red!18}\makecell{$100.0\tabstd{0.0}$ \\ \scriptsize (\$0.39)} & \makecell{$68.2\tabstd{42.9}$ \\ \scriptsize (\$6.31)} & \makecell{$73.9\tabstd{36.9}$ \\ \scriptsize (\$0.01)} & \makecell{$99.4\tabstd{0.4}$ \\ \scriptsize (\$6.96)} & \makecell{$10.0\tabstd{10.3}$ \\ \scriptsize (\$0.14)} & \cellcolor{red!12}\makecell{$76.2\tabstd{13.7}$ \\ \scriptsize (\$0.99)} & \makecell{$61.4\tabstd{23.1}$ \\ \scriptsize (\$1.10)} & \cellcolor{red!6}\makecell{$76.2\tabstd{12.0}$ \\ \scriptsize (\$0.67)} & \makecell{$49.4\tabstd{40.5}$ \\ \scriptsize (\$0.36)} & \makecell{$68.3\tabstd{6.9}$ \\ \scriptsize (\$0.74)} \\
GPT-5.6 terra & \cellcolor{red!18}\makecell{$100.0\tabstd{0.0}$ \\ \scriptsize (\$4.38)} & \cellcolor{red!18}\makecell{$100.0\tabstd{0.0}$ \\ \scriptsize (\$0.27)} & \cellcolor{red!18}\makecell{$100.0\tabstd{0.0}$ \\ \scriptsize (\$5.73)} & \makecell{$99.4\tabstd{0.1}$ \\ \scriptsize \tabpct{49}} & \cellcolor{red!18}\makecell{$100.0\tabstd{0.0}$ \\ \scriptsize (\$0.23)} & \makecell{$99.3\tabstd{0.0}$ \\ \scriptsize \tabpct{44}} & \makecell{$23.3\tabstd{13.9}$ \\ \scriptsize (\$1.74)} & \makecell{$70.9\tabstd{18.1}$ \\ \scriptsize (\$6.99)} & \cellcolor{red!24}\makecell{$83.1\tabstd{6.0}$ \\ \scriptsize (\$14.96)} & \makecell{$71.4\tabstd{9.1}$ \\ \scriptsize (\$5.05)} & \makecell{$59.5\tabstd{15.7}$ \\ \scriptsize (\$5.12)} & \makecell{$64.2\tabstd{27.9}$ \\ \scriptsize \tabpct{73}} \\
Gemini-3.5-flash-lite & \makecell{$95.4\tabstd{8.0}$ \\ \scriptsize (\$1.82)} & \makecell{$67.8\tabstd{55.7}$ \\ \scriptsize (\$0.50)} & \cellcolor{red!18}\makecell{$100.0\tabstd{0.0}$ \\ \scriptsize (\$0.81)} & \makecell{$57.0\tabstd{34.3}$ \\ \scriptsize (\$15.10)} & \makecell{$99.8\tabstd{0.1}$ \\ \scriptsize (\$5.15)} & \makecell{$37.1\tabstd{27.9}$ \\ \scriptsize (\$6.72)} & \makecell{$0.0\tabstd{0.0}$ \\ \scriptsize (\$0.01)} & \makecell{$0.0\tabstd{0.0}$ \\ \scriptsize (\$0.01)} & \makecell{$0.0\tabstd{0.0}$ \\ \scriptsize (\$3.07)} & \makecell{$0.0\tabstd{0.0}$ \\ \scriptsize (\$0.08)} & \makecell{$0.0\tabstd{0.0}$ \\ \scriptsize (\$0.04)} & \makecell{$2.5\tabstd{4.3}$ \\ \scriptsize (\$0.36)} \\
Gemini-3.5-flash & \cellcolor{red!18}\makecell{$100.0\tabstd{0.0}$ \\ \scriptsize (\$5.61)} & \cellcolor{red!18}\makecell{$100.0\tabstd{0.0}$ \\ \scriptsize (\$1.26)} & \makecell{$99.9\tabstd{0.2}$ \\ \scriptsize (\$5.78)} & \makecell{$97.6\tabstd{1.0}$ \\ \scriptsize \tabpct{14}} & \cellcolor{red!18}\makecell{$100.0\tabstd{0.0}$ \\ \scriptsize (\$0.74)} & \makecell{$96.5\tabstd{4.7}$ \\ \scriptsize \tabpct{14}} & \makecell{$35.4\tabstd{56.0}$ \\ \scriptsize (\$8.53)} & \makecell{$50.8\tabstd{44.1}$ \\ \scriptsize \tabpct{69}} & \makecell{$48.1\tabstd{24.7}$ \\ \scriptsize \tabpct{45}} & \makecell{$65.8\tabstd{29.1}$ \\ \scriptsize \tabpct{71}} & \makecell{$21.7\tabstd{37.5}$ \\ \scriptsize \tabpct{92}} & \makecell{$54.4\tabstd{40.8}$ \\ \scriptsize \tabpct{51}} \\
Claude-haiku-4.5 & \makecell{$42.4\tabstd{23.6}$ \\ \scriptsize (\$6.39)} & \makecell{$81.7\tabstd{28.7}$ \\ \scriptsize (\$7.28)} & \makecell{$95.6\tabstd{3.5}$ \\ \scriptsize (\$8.07)} & \makecell{$35.2\tabstd{35.3}$ \\ \scriptsize \tabpct{72}} & \makecell{$88.8\tabstd{19.3}$ \\ \scriptsize \tabpct{84}} & \makecell{$96.7\tabstd{1.7}$ \\ \scriptsize \tabpct{12}} & \makecell{$7.9\tabstd{7.3}$ \\ \scriptsize (\$2.29)} & \makecell{$21.2\tabstd{22.9}$ \\ \scriptsize (\$2.26)} & \makecell{$1.6\tabstd{1.6}$ \\ \scriptsize (\$5.91)} & \makecell{$28.6\tabstd{9.1}$ \\ \scriptsize (\$1.18)} & \makecell{$32.0\tabstd{7.6}$ \\ \scriptsize (\$1.25)} & \makecell{$73.8\tabstd{10.6}$ \\ \scriptsize (\$5.57)} \\
Qwen3.7-plus & \cellcolor{red!18}\makecell{$100.0\tabstd{0.0}$ \\ \scriptsize (\$1.07)} & \makecell{$97.6\tabstd{3.9}$ \\ \scriptsize (\$0.88)} & \cellcolor{red!18}\makecell{$100.0\tabstd{0.0}$ \\ \scriptsize (\$0.85)} & \makecell{$89.6\tabstd{13.5}$ \\ \scriptsize (\$17.73)} & \cellcolor{red!18}\makecell{$100.0\tabstd{0.0}$ \\ \scriptsize (\$5.37)} & \makecell{$98.4\tabstd{1.9}$ \\ \scriptsize (\$18.52)} & \makecell{$64.0\tabstd{19.8}$ \\ \scriptsize (\$4.44)} & \makecell{$74.1\tabstd{22.9}$ \\ \scriptsize (\$2.64)} & \cellcolor{red!30}\makecell{$85.2\tabstd{6.4}$ \\ \scriptsize (\$5.53)} & \cellcolor{red!30}\makecell{$90.0\tabstd{3.1}$ \\ \scriptsize (\$5.56)} & \makecell{$65.5\tabstd{38.4}$ \\ \scriptsize (\$2.70)} & \makecell{$71.7\tabstd{17.2}$ \\ \scriptsize \tabpct{72}} \\
MiniMax-M3 & \makecell{$97.7\tabstd{4.0}$ \\ \scriptsize (\$1.54)} & \makecell{$89.7\tabstd{15.0}$ \\ \scriptsize \tabpct{81}} & \cellcolor{red!18}\makecell{$100.0\tabstd{0.0}$ \\ \scriptsize (\$1.11)} & \makecell{$50.2\tabstd{43.8}$ \\ \scriptsize (\$7.42)} & \makecell{$85.1\tabstd{23.2}$ \\ \scriptsize (\$0.51)} & \makecell{$91.3\tabstd{7.9}$ \\ \scriptsize \tabpct{99}} & \makecell{$57.7\tabstd{25.5}$ \\ \scriptsize (\$4.41)} & \makecell{$33.3\tabstd{55.0}$ \\ \scriptsize (\$2.54)} & \cellcolor{red!6}\makecell{$76.2\tabstd{16.7}$ \\ \scriptsize (\$8.89)} & \makecell{$61.1\tabstd{20.7}$ \\ \scriptsize (\$10.75)} & \makecell{$67.5\tabstd{31.5}$ \\ \scriptsize (\$4.16)} & \cellcolor{red!18}\makecell{$83.5\tabstd{11.5}$ \\ \scriptsize \tabpct{87}} \\
DeepSeek-V4-flash & \makecell{$99.9\tabstd{0.2}$ \\ \scriptsize (\$0.67)} & \makecell{$97.7\tabstd{4.0}$ \\ \scriptsize (\$0.23)} & \makecell{$99.9\tabstd{0.2}$ \\ \scriptsize (\$0.47)} & \makecell{$46.8\tabstd{10.9}$ \\ \scriptsize (\$6.32)} & \makecell{$54.7\tabstd{50.7}$ \\ \scriptsize (\$0.05)} & \makecell{$61.6\tabstd{25.7}$ \\ \scriptsize (\$7.61)} & \makecell{$25.9\tabstd{22.6}$ \\ \scriptsize (\$0.66)} & \makecell{$70.4\tabstd{20.7}$ \\ \scriptsize (\$0.87)} & \cellcolor{red!18}\makecell{$82.5\tabstd{11.1}$ \\ \scriptsize (\$3.10)} & \makecell{$39.6\tabstd{46.3}$ \\ \scriptsize (\$0.35)} & \cellcolor{red!12}\makecell{$77.4\tabstd{30.8}$ \\ \scriptsize (\$1.07)} & \cellcolor{red!24}\makecell{$85.1\tabstd{7.1}$ \\ \scriptsize (\$3.17)} \\
\bottomrule
\end{tabular}}
\end{table}

\section{Online Harness Optimization}
\label{apdx:online_harness}

We test whether a coding agent can improve a long-lived agent's harness while the agent is running, in the spirit of end-to-end harness optimization~\citep{Lee2026MetaHarnessEO}. The setting is breakout news detection with a GPT-5.6-luna actor under the sleep harness on one four-week episode, cut into nine three-day spans. At every cut, a proposer (a coding agent, OpenAI Codex) reads the actor's trajectory so far together with the hindsight available at that point, \textit{i.e.}, the price breakpoints and settled outcomes up to the cut, and edits the actor's harness code and state in place. All three seeds then resume the next span from the same world checkpoint with the edited harness. The proposer sees no future data, no held-out validation set, and no scorer; it may change anything in the actor program. We run the proposer with two backbones, GPT-5.6-sol and GPT-5.6-luna, and compare against the same harness left unchanged for the whole episode.

% Appendix: online harness optimization on breakout news detection, May episode (source:
% docs/improver_results/reflection_memory_and_online_improver_report_v1.md). Requires \usepackage{wrapfig}.
\begin{table}
\vspace{-\baselineskip}
\centering
\small
\setlength{\tabcolsep}{4pt}
\caption{Online harness updates on breakout news detection (GPT-5.6-luna actor, sleep harness, three seeds): coverage F1 ($\times 100$).}
\label{tab:online-harness}
\begin{tabular}{@{}lccr@{}}
\toprule
Harness & Coverage  \\
\midrule
No update                      & $11.2 \pm 0.7$   \\
Update, GPT-5.6-sol proposer   & $13.3 \pm 3.8$  \\
Update, GPT-5.6-luna proposer  & $7.3 \pm 3.0$  \\
\bottomrule
\end{tabular}
\end{table}

Table~\ref{tab:online-harness} summarizes the outcome. With GPT-5.6-sol as the proposer, the updated harness achieves $13.3$ coverage F1 against $11.2$ for the static harness; in contrast, GPT-5.6-luna proposer (same as the actor model) drops coverage F1 to $7.3$. We notice that GPT-5.6-sol keeps the language model in the loop: it adds a second, stateless LLM review of every proposed notification with per-market confidence thresholds that it recalibrates at each cut from the previous span's accepted and rejected proposals, bounds how long an agent may sleep between news checks, and renders the settled outcomes so far into each agent's prompt. Its notes read like an experimenter's, attributing each missed breakpoint to cadence, direction, or reviewer strictness and adjusting a threshold.
On the other hand, GPT-5.6-luna replaces judgment with fragile rules: a proposed notification must carry a catalyst term from a 45-entry list in its headline, a minimum count of direction-specific words, no speculative modal, per-market identity terms, and must pass a recent-price check, unless it matches one of 47 ``strong title overrides'' that copy the exact headline forms of breakpoints the proposer had already seen. Hard rules distilled from a handful of past events do not transfer to the next three days, leading to performance below no update baseline.

\section{Seed-to-Seed Spread by Trigger Mechanism}
Table~\ref{tab:seed-spread} reports, for each task and trigger mechanism, how far repeated runs of the same model-and-trigger configuration spread apart. The spread changes with the trigger mechanism on popularity prediction, market price check, and the two web tasks, and stays within a few points under every trigger on the two detection tasks.
% Appendix: seed spread by trigger (source: docs/scripted_analysis/seed_spread_by_trigger_v1.py).
\begin{table}[t]
\centering
\small
\setlength{\tabcolsep}{5pt}
\caption{Spread between repeated runs of the same model-and-trigger configuration, by trigger mechanism: pooled within-cell standard deviation of per-seed scores (metric $\times 100$) among the five strongest models per task, the models of Figure~\ref{fig:anova-tape}.}
\label{tab:seed-spread}
\begin{tabular}{@{}lcccc@{}}
\toprule
Task & Sleep & Watcher & Cron & All \\
\midrule
Breakout News Detection & 3.1 & 2.5 & 3.1 & 2.9 \\
Forecast Portfolio & 15.0 & 20.1 & 14.1 & 16.1 \\
Resolution Detection & 3.4 & 3.8 & 3.9 & 3.7 \\
Popularity Prediction & 11.8 & 16.0 & 6.1 & 12.3 \\
Daily Digest & 0.9 & 2.6 & 0.1 & 1.6 \\
Market Price Check & 27.5 & 17.5 & 4.7 & 19.0 \\
Web Filing & 19.3 & 30.0 & 14.3 & 22.2 \\
Web Broker Ops & 26.1 & 30.0 & 20.2 & 25.8 \\
\bottomrule
\end{tabular}
\end{table}

\section{Data Licenses}
\label{apdx:license}
News articles are extracted from Common Crawl CC-NEWS (March to July 2026) under Common Crawl's Terms of Use; article text is used only inside the simulator and is not redistributed. Market metadata and hourly prices are fetched from Polymarket's public Gamma and CLOB APIs. Reddit submissions and comment trees for ten AI subreddits are obtained through the Arctic Shift public archive for non-commercial research, consistent with Reddit's Public Content Policy; no content is used for model training. We release the collection scripts, identifiers, and derived labels rather than raw third-party content.

Financial filings are taken from the SEC's EDGAR public data endpoints\footnote{\url{https://www.sec.gov/search-filings/edgar-application-programming-interfaces}}: the ticker-to-CIK map, each company's submissions index (accession numbers, form types, and acceptance timestamps), and its XBRL company-facts records. These are U.S. government records in the public domain (17 U.S.C.\ \S~105), which the SEC designates as public information that may be copied and distributed; we do not use the filing documents or exhibits themselves, only these structured records.

% Appendix: example of an agent's wait policy under TM-A (ReACT + sleep) on the Fed rate-cut market,
% GPT-5.6 luna, bnpm w10, seed s5, first week. Source: docs/analysis/appendix_wait_policies_bnpm_luna_w10_v1.md
% (generated by docs/scripted_analysis/appendix_wait_policies_bnpm_v1.py). Requires: booktabs, tabularx, array.
\begin{table}[t]
\centering
\scriptsize
\setlength{\tabcolsep}{4pt}
\renewcommand{\arraystretch}{1.1}
\caption{GPT-5.6-luna w/ wait harness, in the first week on the market ``Will no Fed rate cuts happen in 2026?'' on breakout news detection. Each row is one \texttt{sleep} call: when the agent woke, when it chose to wake next, whether it filed a claim in that waking, and the thought it attached to the wait.}
\label{tab:wait-tma-luna}
\begin{tabularx}{\textwidth}{@{} l l l >{\raggedright\arraybackslash}X @{}}
\toprule
Wake (UTC) & Sleep until & Claim $\to$ outcome & Final thought \\
\midrule
03-01 00:00 & 03-02 00:00 & up $\to$ false alarm & Wait for the claim to resolve or for new macro news; avoid additional claims while this standing claim is active. \\
03-02 00:00 & 03-02 00:05 & -- & The initial claim remains active through 00:00Z; wait briefly for resolution before considering the fresh Iran-oil catalyst. \\
03-02 00:05 & 03-03 00:00 & \textbf{up $\to$ covered} & Wait for the Iran-oil claim to resolve or expire; avoid locking the market with another claim. \\
\textbf{03-02 13:55} & & & \textbf{Breakpoint $\uparrow$ $+0.024$ (0.102 $\to$ 0.127): covered by the 00:05 claim.} \\
03-03 00:00 & 03-04 00:00 & -- & The March 2 upward move likely resolves the active claim; wait until it is settled before seeking another opportunity. \\
\textbf{03-03 07:01} & & & \textbf{Breakpoint $\uparrow$ $+0.052$ (0.127 $\to$ 0.178): missed, no claim standing.} \\
03-04 00:00 & 03-05 00:00 & -- & March 3 articles describe an already-priced oil-driven repricing, not a new pre-breakout catalyst; avoid a late claim and wait for fresh news. \\
03-05 00:00 & 03-05 06:00 & -- & No fresh March 5 Fed catalyst is visible yet, and the late March 4 drift is not an identifiable new breakout; wait for the US session/news cycle. \\
03-05 06:00 & 03-06 00:00 & -- & A sharp downside move began around 02:27Z, before Williams coverage published at 02:48Z, so notifying now would be late and scoreless; wait for a fresh catalyst. \\
03-06 00:00 & 03-07 00:00 & up $\to$ wrong direction & Wait for the March 5 oil/inflation claim to resolve or expire before making another market claim. \\
\textbf{03-06 13:41} & & & \textbf{Breakpoint $\downarrow$ $-0.075$ (0.187 $\to$ 0.112): missed; the standing claim pointed the other way.} \\
03-07 00:00 & 03-08 00:00 & -- & The February payrolls shock was published and priced on March 6, while the active upward claim covered that window; a new downside notification now would be late. Wait for fresh news. \\
\bottomrule
\end{tabularx}
\end{table}

\begin{table}[t]
\centering
\scriptsize
\setlength{\tabcolsep}{4pt}
\renewcommand{\arraystretch}{1.1}
\caption{GPT-5.6-luna w/ watcher harness in the first week on the market ``Will no Fed rate cuts happen in 2026?'' on breakout news detection. The agent has no \texttt{sleep} tool; it waits by running programs it writes itself. Each plain row is one waking: the program whose wait ended it (a news watcher handing over a headline, or the blind-wait program \texttt{sleep.py} reaching its wake time).}
\label{tab:wait-tmb-luna}
\begin{tabularx}{\textwidth}{@{} l l l l >{\raggedright\arraybackslash}X @{}}
\toprule
Wake (UTC) & Program run & File written & Claim $\to$ outcome & Final thought \\
\midrule
03-01 00:00 & \texttt{fed\_watch.py} & write \texttt{fed\_watch.py} & -- & Dry-run the 30-minute Fed-news watcher before advancing simulated time. \\
03-01 00:00 & \texttt{sleep.py} & edit \texttt{sleep.py} & up $\to$ false alarm & Advance 30 minutes while the standing claim remains unresolved. \\
03-01 00:30 & \texttt{sleep.py} & edit \texttt{sleep.py} & -- & Advance to just beyond the standing claim expiration. \\
03-02 00:05 & \texttt{sleep.py} & edit \texttt{sleep.py} & \textbf{up $\to$ covered} & Advance beyond the 24-hour resolution window. \\
\textbf{03-02 13:55} & & & & \textbf{Breakpoint $\uparrow$ $+0.024$ (0.102 $\to$ 0.127): covered by the 00:05 claim.} \\
03-03 00:10 & \texttt{fed\_watch2.py} & write \texttt{fed\_watch2.py} & -- & Continue polling for a US-specific Fed catalyst; no credible article is present now. \\
03-03 00:40 & \texttt{fed\_watch2.py} & -- & -- & The conflict article was published after the visible move, so it cannot support a timely claim; continue polling for a genuinely new catalyst. \\
03-03 01:10 & \texttt{sleep.py} & edit \texttt{sleep.py} & \textbf{up $\to$ covered} & Advance to after expiration of the latest standing claim. \\
\textbf{03-03 07:01} & & & & \textbf{Breakpoint $\uparrow$ $+0.052$ (0.127 $\to$ 0.178): covered by the 01:10 claim.} \\
03-04 01:15 & \texttt{fed\_watch2.py} & -- & -- & The market was already declining before the 01:02 article, so it cannot justify a timely claim; keep monitoring without notifying. \\
03-04 01:45 & \texttt{fed\_watch2.py} & -- & -- & The article arrived after the market's decline began and does not support a fresh upside breakout; continue monitoring without claiming. \\
03-04 02:15 & \texttt{fed\_watch2.py} & -- & -- & The Kashkari article was published after the upside move began at 01:50, so a notification now would be late; continue monitoring for the next breakout catalyst. \\
03-04 03:15 & \texttt{sleep.py} & edit \texttt{sleep.py} & up $\to$ false alarm & Advance beyond the standing claim window. \\
03-05 03:20 & \texttt{fed\_watch2.py} & -- & -- & The ISM article supports no cuts, but the 02:27 move was sharply downward, so an upside claim would be wrong-direction and late; continue monitoring for the next breakout. \\
03-05 04:20 & \texttt{fed\_watch2.py} & -- & -- & No direct Fed-policy catalyst appeared after the decline; continue automated monitoring. \\
03-05 04:50 & \texttt{sleep.py} & edit \texttt{sleep.py} ($\times$2) & -- & Advance to a later checkpoint for focused monitoring. \\
03-06 05:00 & \texttt{sleep.py} & edit \texttt{sleep.py} & -- & Advance to the jobs-data window without spending on low-relevance news. \\
03-06 12:00 & \texttt{sleep.py} & edit \texttt{sleep.py} & \textbf{down $\to$ covered} & Advance beyond the current standing claim window. \\
\textbf{03-06 13:41} & & & & \textbf{Breakpoint $\downarrow$ $-0.075$ (0.187 $\to$ 0.112): covered by the 12:00 claim on the payrolls report.} \\
03-07 12:05 & \texttt{fed\_watch2.py} & -- & -- & No new direct Fed-policy story is visible after the payroll-driven moves; continue automated monitoring. \\
\bottomrule
\end{tabularx}
\end{table}

\begin{table}[t]
\centering
\scriptsize
\setlength{\tabcolsep}{4pt}
\renewcommand{\arraystretch}{1.1}
\caption{GPT-5.6-luna w/ cron harness in the first week on the market ``Will no Fed rate cuts happen in 2026?'' on breakout news detection. A fixed cron wakes the agent every six hours. The agent can create one-shot or recurring schedules of its own, but it never does in this run.}
\label{tab:wait-tmd-luna}
\begin{tabularx}{\textwidth}{@{} l >{\raggedright\arraybackslash}p{2.7cm} l >{\raggedright\arraybackslash}X @{}}
\toprule
Trigger (UTC) & Tools called & Claim $\to$ outcome & Final thought \\
\midrule
03-01 00:00 (bootstrap) & search $\times$5,  article,  notify & up $\to$ false alarm & No additional action is appropriate while the standing claim is unresolved. \\
03-01 00:05 & -- & -- & The market is locked by the active claim, so I am monitoring without taking another action. \\
03-01 06:05 & search & -- & The market is locked by the existing claim, so the new corroborating article cannot be cited in another notification. \\
03-01 12:05 & prices,  search & -- & The active claim remains unresolved and prevents replacement or additional notification. \\
03-01 18:05 & search,  prices & -- & No new notification is allowed while the existing claim remains active, and no definitive breakout is visible. \\
03-02 00:05 & search $\times$2,  prices,  article,  notify $\times$4 & \textbf{up $\to$ covered} & The new standing claim is accepted; no further notification is possible until it resolves. \\
03-02 06:05 & prices,  search & -- & No new notification can be placed until the opposite breakout formally resolves the standing claim. \\
03-02 12:05 & prices,  search,  notify & up (rejected) & One standing claim is still active; no further notification can be placed. \\
\textbf{03-02 13:55} & & & \textbf{Breakpoint $\uparrow$ $+0.024$ (0.102 $\to$ 0.127): covered by the 03-02 00:05 claim.} \\
03-02 18:05 & search & -- & The active claim already covers the upward repricing; further claims are blocked by the one-standing-claim rule. \\
03-03 00:05 & prices,  search $\times$2 & -- & Avoiding a low-confidence claim after an already-covered move and without a clearly imminent fresh breakout. \\
03-03 06:05 & prices,  search & -- & Avoiding a low-confidence claim when price action is gradual rather than breakout-like. \\
\textbf{03-03 07:01} & & & \textbf{Breakpoint $\uparrow$ $+0.052$ (0.127 $\to$ 0.178): missed, no claim standing.} \\
03-03 12:05 & search,  prices & -- & The visible move is already underway, so this is not actionable under the pre-breakout requirement. \\
03-03 18:05 & search,  prices & -- & The likely move is already underway and directionally conflicted with later news, so a new claim would be late and low-confidence. \\
\multicolumn{4}{c}{$\vdots$} \\
03-06 06:05 & prices,  search & -- & The standing claim already covers the upward thesis and the market remains locked. \\
03-06 12:05 & prices,  search,  notify & up $\to$ wrong direction & The claim is accepted and the market is locked until resolution. \\
\textbf{03-06 13:41} & & & \textbf{Breakpoint $\downarrow$ $-0.075$ (0.187 $\to$ 0.112): missed; the claim filed 96 minutes earlier pointed up.} \\
03-06 18:05 & prices,  search & -- & The market is locked by the standing claim, and the latest news supports cuts rather than a new upward breakout. \\
03-07 00:05 & prices,  search & -- & The current standing claim already occupies the market and latest evidence does not support a fresh upward breakout. \\
\bottomrule
\end{tabularx}
\end{table}

\begin{table}[t]
\centering
\scriptsize
\setlength{\tabcolsep}{4pt}
\renewcommand{\arraystretch}{1.1}
\caption{GPT-5.6-terra w/ wait harness on breakout news detection for the market ``Will no Fed rate cuts happen in 2026?''. The agent anticipates payroll release by adaptively scheduling its wake. Out of 6 repeated runs of GPT-5.6-luna and terra, this anticipitation of the payroll release is observed on 1 seed on GPT-5.6-terra w/ wait harness, 2 seeds on GPT-5.6-terra w/ watcher harness, 1 seeds on GPT-5.6-luna w/ watcher harness, and 0 seed on the cron harness.
}
\label{tab:wait-tma-terra}
\begin{tabularx}{\textwidth}{@{} l l l >{\raggedright\arraybackslash}X @{}}
\toprule
Wake (UTC) & Sleep until & Claim $\to$ outcome & Final thought \\
\midrule
03-06 00:00 & 03-06 13:40 & -- & Wait for the imminent US employment release, which is more likely than overnight commentary to create a discrete Fed-path breakout. \\
03-06 13:40 & 03-06 18:00 & \textbf{down $\to$ covered} & Standing down claim is active; wait while the jobs shock is incorporated. \\
\textbf{03-06 13:41} & & & \textbf{Breakpoint $\downarrow$ $-0.075$ (0.187 $\to$ 0.112): covered by the 13:40 claim on the payrolls report.} \\
03-06 18:00 & 03-07 12:00 & -- & No distinct post-payroll catalyst appears; preserve optionality and resume on the next news cycle. \\
\bottomrule
\end{tabularx}
\end{table}

% Appendix: the news watcher a TM-B agent wrote at bootstrap on the Fed rate-cut market (GPT-5.6 luna, bnpm w10,
% seed s1), verbatim from workspace/agents/m-616902/fed_watch.py; copy kept at assets/bnpm_examples/fed_watch_luna.py.
% Requires \usepackage{listings} (and xcolor). No shell-escape needed. If the preamble already has a global
% \lstset, the local \lstset below only overrides inside this float.
\begin{figure}[t]
\lstset{
  language=Python,
  basicstyle=\scriptsize\ttfamily,
  keywordstyle=\bfseries\color{blue!60!black},
  commentstyle=\itshape\color{gray},
  stringstyle=\color{green!40!black},
  showstringspaces=false,
  breaklines=true,
  breakatwhitespace=false,
  postbreak=\mbox{\textcolor{gray}{$\hookrightarrow$}\space},
  columns=fullflexible,
  keepspaces=true,
  numbers=left,
  numberstyle=\tiny\color{gray},
  numbersep=6pt,
  xleftmargin=1.6em,
  frame=single,
  framerule=0.3pt,
  rulecolor=\color{gray!60},
  aboveskip=0pt,
  belowskip=4pt,
  literate={->}{{$\rightarrow$}}2 {—}{{\textemdash}}1,
}
\begin{lstlisting}
"""example_gatekeeper.py — a minimal, runnable wake-condition program:
poll news on a cadence, hand over fresh keyword-matching articles.

Copy and edit it (write_file / edit_file), dry-run with
run_program(path, validate=true), then run_program(path). Wake times
live in code — wait(now() + interval); run_program's `until` is only a
hard timeout backstop. State persists in workspace files, so a trigger
interrupting the run loses nothing: just re-run the same program.

NB: `envkit` exists only inside your run_program workspace — this file is
a template, not an importable module of the repo.
"""

import json
from datetime import timedelta

from envkit import handover, now, search_news, wait, workspace

QUERY = '"rate cut" OR "FOMC"'      # BM25: quoted phrases, AND/OR
KEYWORDS = ("cut", "hike", "fed")   # cheap title filter on top of BM25
POLL_EVERY = timedelta(minutes=30)  # cadence lives in code

seen_file = workspace() / "gatekeeper_seen.json"
seen = set(json.loads(seen_file.read_text())) if seen_file.exists() else set()

while True:
    t = now()
    page = search_news(
        q=QUERY,
        date_from=(t - POLL_EVERY).strftime("%Y-%m-%dT%H:%M:%SZ"))
    hits = [h for h in page["results"]
            if h["news_id"] not in seen
            and any(k in h["title"].lower() for k in KEYWORDS)]
    seen |= {h["news_id"] for h in page["results"]}
    seen_file.write_text(json.dumps(sorted(seen)))
    if hits:
        handover({"matched": [
            {"news_id": h["news_id"], "title": h["title"],
             "published": h["published"]} for h in hits[:5]]})
    wait(t + POLL_EVERY)
\end{lstlisting}
\caption{Default watcher program shipped together with a \benchmark{} task for the watcher harness, exemplified with the breakout news detection task.}
\label{lst:example-gatekeeper}
\end{figure}

% Appendix: the news watcher a TM-B agent wrote at bootstrap on the Fed rate-cut market (GPT-5.6 luna, bnpm w10,
% seed s1), verbatim from workspace/agents/m-616902/fed_watch.py; copy kept at assets/bnpm_examples/fed_watch_luna.py.
% Requires \usepackage{listings} (and xcolor). No shell-escape needed. If the preamble already has a global
% \lstset, the local \lstset below only overrides inside this float.
\begin{figure}[t]
\lstset{
  language=Python,
  basicstyle=\scriptsize\ttfamily,
  keywordstyle=\bfseries\color{blue!60!black},
  commentstyle=\itshape\color{gray},
  stringstyle=\color{green!40!black},
  showstringspaces=false,
  breaklines=true,
  breakatwhitespace=false,
  postbreak=\mbox{\textcolor{gray}{$\hookrightarrow$}\space},
  columns=fullflexible,
  keepspaces=true,
  numbers=left,
  numberstyle=\tiny\color{gray},
  numbersep=6pt,
  xleftmargin=1.6em,
  frame=single,
  framerule=0.3pt,
  rulecolor=\color{gray!60},
  aboveskip=0pt,
  belowskip=4pt,
  literate={->}{{$\rightarrow$}}2,
}
\begin{lstlisting}
import json
from datetime import timedelta
from envkit import handover, now, search_news, wait, workspace

POLL = timedelta(minutes=30)
FILE = workspace() / 'fed_seen.json'
seen = set(json.loads(FILE.read_text())) if FILE.exists() else set()
# Broad retrieval, then strict title filter to avoid spending claims on noise.
QUERY = '("Federal Reserve" OR Fed OR FOMC OR Powell OR "interest rate" OR "rate cut" OR "rate hike" OR inflation)'
TERMS = ('fed','federal reserve','fomc','powell','rate cut','rate hike','interest rate','inflation','basis points','monetary policy')
while True:
    t = now()
    page = search_news(q=QUERY, date_from=(t-POLL).strftime('%Y-%m-%dT%H:%M:%SZ'), date_to=t.strftime('%Y-%m-%dT%H:%M:%SZ'))
    fresh=[]
    for h in page.get('results',[]):
        nid=h.get('news_id'); title=h.get('title','')
        if nid not in seen and any(x in title.lower() for x in TERMS):
            fresh.append({'news_id':nid,'title':title,'published':h.get('published','')})
        if nid: seen.add(nid)
    FILE.write_text(json.dumps(sorted(seen)))
    if fresh:
        handover({'market_id':'616902','matched':fresh[:10], 'checked_at':t.isoformat()})
    wait(t+POLL)
\end{lstlisting}
\caption{The watcher program authored by GPT-5.6-luna w/ watcher harness agent for the market ``Will no Fed rate cuts happen in 2026?''. The program polls the news index every 30 minutes, hands over to the agent only when a fresh headline matches its keyword filter, and persists the set of seen articles across runs. The program is reused and replaced by \texttt{fed\_watcher2.py} in later during the run, as in the agent trace shown in Table~\ref{tab:wait-tmb-luna}.}
\label{lst:fed-watch-luna}
\end{figure}

% Appendix: the raw task instruction for popularity prediction (tasks/reddit_ai_popularity/INSTRUCTION.md), verbatim;
% copy kept at assets/tasks/instruction_reddit_ai_popularity.md. `${...}` are the constructor's placeholders,
% filled per cell at provisioning (the rendered copy each agent saw is workspace/INSTRUCTION.md in its run dir).
% Requires \usepackage{listings} (and xcolor); pdflatex with UTF-8 input is enough — the non-ASCII glyphs in the
% file (≥ ≤ − — – … Σ) are mapped to math symbols via `literate` below, no fontspec needed.
% At \footnotesize the 85 lines overrun one page: trim lines in the .md copy, or switch basicstyle to \scriptsize.
\begin{figure}[p]
\lstset{
  language={},
  basicstyle=\scriptsize\ttfamily,
  columns=fullflexible,
  keepspaces=true,
  breaklines=true,
  breakatwhitespace=true,
  postbreak=\mbox{\textcolor{gray}{$\hookrightarrow$}\space},
  showstringspaces=false,
  frame=single,
  framerule=0.3pt,
  rulecolor=\color{gray!60},
  xleftmargin=0.5em,
  aboveskip=0pt,
  belowskip=4pt,
  extendedchars=true,
  inputencoding=utf8,
  literate=%
    {≥}{{$\geq$}}1
    {≤}{{$\leq$}}1
    {−}{{$-$}}1
    {—}{{---}}2
    {–}{{--}}1
    {…}{{\dots}}1
    {Σ}{{$\Sigma$}}1,
}
\begin{lstlisting}
# Task: recommend the AI-subreddit threads that will become popular

You are a long-running automation agent watching a set of AI subreddits (OpenAI,
Anthropic, LocalLLaMA, r/singularity, …). New posts arrive continuously. Your
job: **recommend the posts whose discussion will turn out large — as early as
possible**. A recommendation names one post: payload `{"root_id": ...}` to the
environment's `/notify`.

## Objective

A post's popularity is the size of its comment cascade `${horizon_hours}` hours
after posting: `descendants` = the number of comments in that window. It is
hidden until the reveal, then frozen. A post is a **tail post** when its final
`descendants ≥ ${tail_min_desc}` — those are the posts you are meant to catch.

Your score, computed at the end of the run, is **time-weighted tail recall**:

    TWR = Σ over tail posts of weight(your first recommendation) / (number of tail posts)
    weight = max(0, 1 − (hours between posting and your recommendation) / ${decay_hours})

- Recommend at the moment of posting: weight 1 — full credit.
- The weight falls linearly with delay. Concretely: ${decay_examples}.
- Tail posts you never recommend earn 0. Higher TWR is better.

You get at most **${daily_cap} accepted recommendations per rolling 24 h** —
far fewer than the tail posts that exist — so every slot spent on a post that
fizzles is a tail post forgone. Roughly 3–4% of posts end tail-sized; the
median post ends with only a handful of comments.

## Budget and rate limit

You have **${budget_usd} total for the entire run**.${domain_caps} LLM calls bill real token
cost against it (${llm_price_table}); ...(truncated)

## Rules (violating requests are rejected free of charge, HTTP 400)

- At most ${daily_cap} accepted recommendations per rolling 24 h window.
- Each post can be recommended at most once.
- Only posts made after the run started and not yet revealed are recommendable:
  pre-run history and already-revealed posts are rejected.

## Data visibility

- A post's static metadata (subreddit, author, title, selftext, url, flair) is
  visible from the moment it is posted.
- **The comment cascade is your early signal.** `get_cascade` returns the reply
  tree *as it has grown so far* — every comment with `created_utc ≤ now`, its
  `parent_id` (the tree structure), `author`, and `body`. You watch reply
  velocity and branching build up and judge whether a post is taking off.
- ...(truncated)

## APIs

${APIS}

Every hour you wait before recommending shrinks the credit linearly — but the
longer you watch a cascade, the better you can tell a trender from a dud that
wastes one of your ${daily_cap} daily slots. Spend your attention where it
changes decisions.
\end{lstlisting}
\caption{The popularity-prediction task instruction, as given to the agent. The instruction states the detailed evaluation metrics, the rules (such as single-recommendation-per-post), and the tool documentations. Tool documentations specific to the tasks and triggering mechanisms are appended after the instruction.}
\label{lst:instruction-reddit-ai-popularity}
\end{figure}

\end{document}